\documentclass[12pt]{article}

\usepackage[a4paper,margin=1in]{geometry}
\usepackage{setspace}
\usepackage{graphicx}
\usepackage{subcaption}
\graphicspath{
    {RSSM_Training_Stats/}
    {Figure/}
}

\usepackage[utf8]{inputenc}
\usepackage[T5]{fontenc}
\usepackage[english]{babel} 
\usepackage{lmodern}
\usepackage{hyperref}

\usepackage{amsmath,amssymb}

\usepackage[ruled,vlined]{algorithm2e}

\usepackage{float}
\usepackage{placeins}
\usepackage{rotating}

\usepackage{titlesec}
\usepackage{titling}
\usepackage{fancyhdr}
\usepackage{indentfirst}
\usepackage{caption}
\usepackage[T5]{fontenc}
\usepackage[english]{babel}

\usepackage[numbers,sort&compress]{natbib}
\usepackage{csquotes}

\pretitle{\centering\bfseries\fontsize{14}{20}\selectfont}
\posttitle{}
\preauthor{\centering\small\itshape}
\postauthor{\vspace{1em}}
\predate{}
\postdate{}

\title{%
HanoiWorld :  A Joint Embedding Predictive Architecture Based World Model for Autonomous Vehicle Controller%
}

\author{%
\\[0.5cm]
Tran Tien Dat\textsuperscript{1,2}, Nguyen Hai An\textsuperscript{1,2}, Nguyen Khanh Viet Dung\textsuperscript{1,2}, Nguyen Duy Duc\textsuperscript{1,2}\\[0.5em]
{\small
\textsuperscript{1}Faculty of Mathematics and Informatics, Hanoi University of Science and Technology, Vietnam\\[0.3em]
\textsuperscript{2}Troy University, United States}\\[0.8em]
\centerline{\ttfamily
\{Dat.TT228006, An.NH227993\}@sis.hust.edu.vn}
\centerline{\ttfamily
\{Dung.NKV207947, Duc.ND218129\}@sis.hust.edu.vn}
\centerline{\ttfamily dtran220296@troy.edu}
}

\date{} 

\begin{document}

\begin{samepage}
\maketitle
\end{samepage}

\onehalfspacing

\begin{abstract}
Current attempts of Reinforcement Learning for Autonomous Controller are data-demanding while the result are under-performed, unstable and unable to grapple and anchoring on the concept of safety, and over-concentrate on noise feature dues to the nature of pixel reconstruction. While current Self-Supervised Learning approachs that learning on high-dimensional representation by leveraging the Joint Embedding Predictive Architecture (JEPA) is interesting and effective alternative, as the idea is mimicking the natural of human's brain in acquiring new skill using imagination and minimal sample of observations. This study introduces HanoiWorld, a JEPA-based world model that using recurrent neural network (RNN) for making longterm horizontal planning with effective inference time. Experiments conducted on Highway-Env package with difference enviroment showcase the effective capability of making driving plan while safety-awareness with considerable collision rate in comparison with SOTA baselines.
\end{abstract}

\section{Introduction}
\label{sec:introduction}

Since the first experiment on an Autonomous Vehicle (AV) was conducted in 1986 at Carnegie Mellon University \cite{hacohen_2022_autonomous}, the research domain of developing self-driving vehicles has made significant progress, both in technical and practical applications in the real world. The vehicles are expected to operating safetly while handling the challenges of uncertainty, partial observability, and multi-agent enviroment (interaction between the ego-vehicle and the surroundings vehicles, and obstacle as pedestrians, etc.) \cite{jia_2023_adriveri, zhang_2024_bevworld}. However, as \cite{dulacarnold_2019_challenges} suggest, these challenges does limit the feasibility on deploying and experimenting on such reinforced-learning based controller, and prior work only based on naive transfering paradigms from the physical-enviroment which lead to training instability dues to noise, and data-fragmentation \cite{dulacarnold_2019_challenges, Jia2025, Chen2025}.

From the technical perspective, the decision of AV have traditionally relied on simmulator for experience rollout, and the planning algorithms with reinforcement learning with the assumption of data-abundance as \cite{hessel_2017_rainbow, hoel_2019_combining}, whereas, the classical attempts - such as Monte Carlo Tree Search (MCTS) or belief-space planning under partially observable Markov decision processes (POMDPs) require massive computational overhead within the simmulator for experiment rolling-out without doing the policy training \cite{han_2024_a, morimura_2024_policy, huang_2024_learning}, while these method does offering for resolving the uncertainty, but the scalability is limited \cite{momchil_2025_treeirl}. Additionally such rolling-out strategy tend to amplifying the error over-the-longtime horizon, and lead to the raise of the model inaccuracy \cite{robine_2020_smaller, xiao_2019_learning}. Additionally, the observation-level prediction tends to priortize the visual, and kinematic fidelity as the reconstruction challenge, which may not truely encapsulated the decision-relevant manifold, and leading to inefficencies in object control \cite{sobal_2022_separating, Assran2025, LeCun2022}.

Inspiration from human's capability of acquiring new skill as driving by leveraging the capability of imaginary on the plausible future scenario based on the current interaction with the enviroment - the affordance based theory, and human memory \cite{ha_2018_world, gibson_1979_the, LeCun2022}, which can be formalized into the model-designing implementation as using representation using the self-supervised learning paradigm for learning the enviorment dynamicity. Joint-Embedding Predictive Architectures (JEPA) propose learning latent spaces by directly predicting future representations, without reconstructing raw observations, but only structure alignment between encoder and enforcing the information variability, but not stochasticity on noise for preventing the embedding collaspe as \cite{bardes_2022_vicreg}. State-of-the-art model as V-JEPA-2 extends the idea to large-scale video data for learning the action dynamicity in the yeilded representation from the passive-observation, which can be used later for producing training signal for lightweight action-conditioned controller \cite{Assran2025}. In parallel, the  recurrent state-space models (RSSMs) have been shown to provide an effective mechanism for maintaining compact latent memories that approximate Markovian dynamics under partial observability with the capability of long-term planning using minimal ammount of representation \cite{hafner_2019_learning, Hafner2023}. The attempts of using encoder with long-term RSSM planning model show-case the efficiency and overhead-minimalizing with increasing in training utility than MCTS based approach, while still yeild out sensory grouded action from agents.

Based on these aforementioned works, this result argue that \textit{world-model designing can be potential benefit from the high-quality self-supervised learning embedding from pre-trained encoder as V-JEPA 2 and combine with the usage of long-term planner which can reduce and minimalize the cost of inference while remaining accuracy, and tunable model driving quality}.

The contribution of this studies include 4 keys essential contributions as follow:

\begin{itemize}
    \item A unified perspective on world-model design for autonomous vehicles that emphasizes predictive, representation-level modeling over observation-level simulation which is called HanoiWorld.
    \item Suggesting a JEPA-based encoding strategy, inspired by V-JEPA-2 fine-tuning, for learning decision-relevant latent representations from large-scale video data.
    \item The integration of an RSSM-based latent memory to support approximate Markovian state transitions under partial observability.
    \item A demonstration that a simple MLP-based actor--critic controller can be trained effectively within the learned latent world model, avoiding expensive planning algorithms and complex policy architectures.
\end{itemize}

The rest of the paper shall be showcased as follows; Section \ref{sec:related_work} focusing on the related works and provide the whole conceptual and theoretical foundation on the challenges and related solution; Section \ref{sec:architecture} discusses the suggested proposed world-model designing; Section \ref{sec:experiment} will attempts to provide the experiment description, the usecase and result discussion. Finally, the report shall be conclude on Section \ref{sec:conclusion}.

Additionally, we will release the experimental codebase to facilitate reproducibility at
\href{https://github.com/CS-3331-Fundamental-of-AI/WorldModel-Self-Driving-Car.git}{HANOI--WORLD codebase}.
\section{Related Works}
\label{sec:related_work}

World Model have been proposed as the novel-approach as the novel solution for training the reinforcement learning based controller dues to the unreliability, excessiveness, and inefficent of the model that based on real-world interaction as \cite{dulacarnold_2019_challenges} suggested, while the vehicle constantly work under occlusion with unlimited knowledge on the world trigger the problems of inevitable uncertainty. These challenges co-align and trigger the need of reframe the training approach for the reinforcement-learning controller using the compact while semantic vivid representation for attaining the scalability and reliability  that inspired on human's biological mechanism on learning using affordance and imaginary \cite{LeCun2022, gibson_1979_the}.

\subsection{From Model-Based Reinforcement Learning to Driving World Models}

The conceptual roots of driving world models lie in model-based reinforcement learning (MBRL), where an agent learns a dynamics model and then plans by simulating futures. Progress in MBRL was enabled by shared tooling and standardized benchmarks. OpenAI Gym \cite{Brockman2016} made it easy to compare algorithms across tasks via a uniform API, while the DeepMind Control Suite \cite{Tassa2018} provided a curated set of continuous-control environments that encouraged rigorous evaluation of learning and control. These infrastructures fostered iterative improvements in learning dynamics models, representation learning, and planning.

Modern latent-dynamics agents exemplify the “world model as an imagination engine” viewpoint. DreamerV3 \cite{Hafner2023} demonstrates strong and stable performance across a wide range of environments by learning a compact latent state and optimizing behavior via imagined rollouts. Although most Dreamer-style results are reported outside real driving, the core design principles—predictive latent state, stochastic dynamics, and planning or policy improvement through imagined trajectories—strongly influence driving world model designs. Complementary perspectives on embodied agent design and generalization in RL emphasize that robustness and scalable training protocols matter as much as raw model capacity \cite{Hansen2023}. In autonomous driving, these principles interact with additional constraints: safety, distribution shift, long-horizon decision-making, and multi-agent interactions.

A growing line of work argues that the value of a world model should be judged by downstream utility (e.g., improved planning or safer decisions) rather than by generative fidelity alone. Planning-centric views highlight that an agent can exploit imperfections in a learned model, producing “good” rollouts that do not correspond to the real world. Analyses of embodied world models stress safety as a first-class concern and call for evaluation protocols that expose failure modes, especially those that emerge only in closed-loop control \cite{Baraldi2025}. These concerns become acute in autonomous driving, where rare events dominate risk and a small modeling error can cascade into catastrophic outcomes.

\subsection{Self-Supervised and Predictive Representation Learning for Driving}

Autonomous driving provides abundant unlabeled sensor streams but comparatively limited dense annotations, motivating self-supervised learning (SSL) as a foundation for world models. In vision, SSL matured from contrastive and clustering-based approaches to predictive and distillation-based schemes. DINO \cite{Caron2021} showed that self-distillation without labels can learn semantically meaningful features, and such ideas have inspired driving-specific pretraining efforts that seek transferable representations across time, viewpoint, and weather.

However, driving data introduces distinct pitfalls for generic SSL. Contrastive learning requires defining “positive pairs” that represent the same underlying content under augmentations; in driving scenes with many objects and rapid ego-motion, naive augmentations can destroy correspondence and lead to negative transfer. Generative SSL that reconstructs masked inputs can be expensive for 3D data and may force the model to predict arbitrary surface details rather than planning-relevant semantics. Several recent works therefore advocate embedding-level prediction and variance regularization as alternatives to either contrastive pairs or explicit reconstruction \cite{Minh2022,Min2024,Zhu2025}.

A particularly influential conceptual framework is the Joint Embedding Predictive Architecture (JEPA) viewpoint, which proposes learning representations by predicting the embeddings of unknown parts of the input given the known parts, rather than reconstructing pixels or using negative pairs \cite{LeCun2022}. Driving provides a compelling application: masked regions in LiDAR or camera space may correspond to multiple plausible surfaces, yet the semantics (e.g., “rear of a car,” “free space behind a truck”) can remain stable in an embedding space. JEPA-style methods can therefore better align with the uncertainty intrinsic to partial observability. For example, JEPA-based LiDAR pretraining predicts BEV embeddings for masked regions and uses explicit variance regularization to prevent representation collapse, yielding consistent gains in downstream 3D detection while reducing pretraining compute relative to dense reconstruction \cite{Zhu2025}.

World models also benefit from discrete or structured latent spaces that stabilize learning and improve sample efficiency. Vector-quantized representations provide one path, but codebook collapse can limit capacity. Online codebook learning strategies such as clustering-based VQ updates aim to keep all codevectors active, improving utilization and reconstruction/generation quality \cite{ZhengVedaldi2023}. In driving, discretized latents may improve controllability, support efficient rollouts, and provide a bridge between geometric and semantic factors.

\subsection{Spatial World States: BEV, Occupancy, and Geometric Abstractions}

Many driving world models adopt spatially grounded world states rather than purely abstract latent vectors. BEV representations offer a convenient coordinate frame that aligns with planning: it naturally represents lanes, drivable space, and other agents, and it facilitates sensor fusion. BEV representations also reduce the burden of viewpoint variation, enabling models to focus on dynamics rather than perspective transformations. Consequently, BEV features are widely used as intermediate states for both perception and prediction, and they serve as a natural substrate for world modeling \cite{Li2022,Li2024}.

Occupancy-based representations extend BEV by modeling 3D free space and occlusion, which are critical for safety. A world model that predicts occupancy can support collision checking, visibility reasoning, and planning under uncertainty. Recent LiDAR-oriented world models stress that camera-only generation may produce visually plausible but geometrically inconsistent futures, whereas occupancy prediction can enforce physical constraints and preserve 3D structure \cite{Bogdoll2025,Li2025}. These works highlight the importance of representing not just objects but also empty space, since the absence of obstacles is as planning-relevant as their presence.

Geometric abstractions are also closely tied to mapping and scene priors. High-definition maps encode lane topology, boundaries, and crosswalks, and several world modeling pipelines treat maps as part of the world state, either as conditioning signals for generation or as latent factors to be predicted. Methods that jointly reason about agent trajectories and map structure aim to ensure that generated futures obey road geometry and traffic rules \cite{Zhao2024,WangDriveDreamer2023}. In addition, surveys of “physical world models” emphasize that effective world models should capture not only statistical regularities but also physically grounded structure and causal relations, particularly when extrapolating beyond the training distribution \cite{Chen2025}.

\subsection{Transformers, Attention, and Interaction-Centric Modeling}

Transformers and attention mechanisms have become central to autonomous driving because they support long-range dependencies and flexible fusion across heterogeneous inputs. In perception and prediction, attention helps focus computation on the most relevant actors and regions of the scene, and it provides a natural way to model interactions among agents. Transformer-based architectures are therefore widely used in modern driving world models, especially when combining multi-view images, point clouds, and map features \cite{Vasudevan2024}.

Interaction-centric modeling is essential because other agents react to each other and to the ego vehicle. World models that treat agents as independent can systematically fail in dense traffic, merges, intersections, and other interactive contexts. Recent work emphasizes representations that capture agent-agent coupling, intent, and right-of-way, often using attention or graph-style message passing to represent joint futures \cite{Long2025,Feng2025}. These ideas are consistent with broader trends in embodied learning, where the world model must represent not only passive dynamics but also the consequences of actions and the strategic responses of others.

Transformers are also influential in self-supervised pretraining and in building scalable “foundation-style” models that transfer across tasks. Papers exploring large-scale training regimes for world models argue that representation, prediction, and planning can be co-trained when the model is sufficiently expressive and the training data is diverse \cite{Wu2025,Xie2025}. This perspective connects to the wider discussion of how to unify perception and control under a single learned model, and it motivates architectures that can condition on both sensory history and action sequences.

\subsection{Multi-Agent and V2X World Models: Cooperative Perception to Cooperative Prediction}

Autonomous driving is inherently multi-agent, and connected autonomy introduces additional channels of information via V2X communication. Cooperative perception is an early instance of “distributed world modeling,” where multiple vehicles or infrastructure sensors share data or features to reduce occlusion and extend sensing range. Public benchmarks for cooperative perception have enabled systematic evaluation of fusion strategies, including early, late, and intermediate fusion \cite{Xu2021}. Intermediate feature sharing is often favored as a balance between accuracy and bandwidth, and it has motivated learned fusion modules that can tolerate localization noise and intermittent communication.

Transformer-based fusion architectures extend cooperative perception by enabling attention over agents and multi-scale features. V2X-focused transformers incorporate mechanisms to handle pose uncertainty, varying sensor modalities, and temporal misalignment due to communication delay \cite{Xu2022}. These settings highlight a key difference between single-agent and multi-agent world models: the “state” is not merely what the ego sees, but a distributed set of partial observations that must be reconciled into a coherent representation. 

Real-world datasets are crucial for validating these ideas because simulation can underrepresent sensor artifacts and the true distribution of road interactions. V2V4Real provides real-world multi-vehicle data with LiDAR, RGB, 3D bounding boxes, and HD maps, designed explicitly for cooperative perception tasks such as cooperative detection, tracking, and sim-to-real adaptation \cite{Xu2023}. Such datasets suggest that future world models for autonomous driving should be designed from the start to support multi-agent fusion and prediction, rather than retrofitting single-agent models to cooperative settings.

Cooperative world models must go beyond “current-state fusion” to “future-state prediction” under shared information. This entails modeling how distributed observations evolve, how agents may act in response to each other, and how communication delays affect belief updates. Works that study cooperative forecasting and interaction in distributed settings argue for robust, uncertainty-aware fusion and for models that degrade gracefully when messages are delayed or missing \cite{Tu2025,Tu2025b}. This line of research also connects to simulators and benchmark environments that can test communication-aware policies under controlled conditions \cite{Audinys2025,Bouzaiene2025}.

\subsection{Safety-Critical Modeling: Accident Prediction, Risk Anticipation, and Vulnerable Road Users}

A primary promise of world models in driving is improved anticipation of rare and safety-critical events. Traditional trajectory prediction benchmarks emphasize average errors on non-critical behavior, which may not correlate with accident risk. Accident prediction benchmarks therefore play an important role in driving world model evaluation. DeepAccident introduces motion and accident prediction in V2X contexts, targeting the challenging setting where the model must predict not only where agents will move, but whether and when a collision will occur and which participants will be involved \cite{WangDeepAccident2023}. Such benchmarks force world models to represent risk factors that may be subtle or long-range, such as occluded cross traffic or rapidly changing gaps.

Risk anticipation also motivates explicit modeling of occlusion and free space, which are critical to collision avoidance. Occupancy-based world models and LiDAR-centric methods emphasize that representing “unknown” regions and visibility is crucial: predicting an agent behind an occluder is a fundamentally different problem from predicting a visible agent. Recent risk-oriented world modeling studies therefore combine geometric priors with predictive uncertainty, aiming to represent multiple plausible futures rather than a single deterministic trajectory \cite{Guan2025,Bogdoll2025}.

Safety also depends on modeling vulnerable road users (VRUs) such as pedestrians and cyclists, especially in dense, mixed traffic. High-resolution trajectory datasets that include VRUs provide essential supervision for interaction-aware world models. OnSiteVRU offers high-density trajectories across complex urban scenarios with fine temporal resolution and contextual information, enabling research on VRU behavior modeling, interaction risk, and safety evaluation \cite{Yan2025}. These datasets complement vehicle-centric benchmarks and highlight that world models must reason about heterogeneous participants with different kinematics, goals, and social norms.

Finally, recent analyses argue that safety evaluation for world models should be multi-dimensional: generative realism is insufficient, and closed-loop planning tests can expose model exploitation or compounding errors. The “safety challenge” perspective calls for stress tests, counterfactual evaluation, and metrics that quantify whether planning with the model improves safety outcomes under distribution shift \cite{Baraldi2025}. This theme strongly motivates research into world models that are not only expressive but also calibrated, robust, and auditable.

\subsection{Generative Simulation: Diffusion, Video World Models, and Language-Conditioned Scenario Generation}

Generative world models have expanded rapidly, driven by diffusion models and advances in controllable video generation. In autonomous driving, video generation is appealing because it can synthesize diverse scenes, including rare events, that are hard to capture in real data. Several works frame driving world modeling as conditional generation of future observations given current context and candidate actions, often incorporating maps, bounding boxes, or trajectories as conditioning signals \cite{Fu2024,Chu2025}. Surveys and methodological papers in this area emphasize that diffusion-based dynamics modeling can produce high-fidelity samples and can incorporate structured priors, but they also note challenges in temporal consistency and action controllability \cite{Baraldi2025,Chen2025}.

DriveDreamer-2 demonstrates a particularly influential direction: integrating language models with world models to enable user-driven simulation. In DriveDreamer-2, an LLM converts user prompts into agent trajectories, a diffusion-based component generates an HD map consistent with those trajectories, and a unified multi-view video generator synthesizes multi-camera driving videos \cite{Zhao2024}. This pipeline aims to generate long-tail scenarios (e.g., abrupt cut-ins) in a user-friendly way and to improve downstream perception training. Related efforts such as DriveDreamer variants emphasize improving cross-view coherence and temporal stability and highlight how structured intermediate representations (trajectories, maps) can improve controllability \cite{WangDriveDreamer2023}.

Generative modeling is also used for counterfactual evaluation and data augmentation in planning-centric contexts. “Dreamer”-style rollouts provide imagined futures in latent space, while diffusion or video models provide rich sensory predictions. Integrative systems such as CarDreamer seek to couple world modeling with downstream decision-making and to use learned imagination as a training signal for driving policies \cite{Gao2024}. Other large-scale training efforts explore how to train world models that support both prediction and planning, often combining self-supervised objectives with policy learning \cite{Assran2025,Wu2025}. Across these works, a recurring open problem is aligning generative fidelity with decision utility: a model can produce realistic-looking videos while still being unreliable for safety-critical planning.

\subsection{Evaluation and Benchmarking: Utility, Generalization, and the Reality Gap}

A persistent difficulty is that there is no single metric that fully captures the quality of a driving world model. Generative metrics as FID - Fréchet Inception Distance, or FVD - Fréchet Video Distance,which can expressed as the Eq.~\eqref{eq:FID} and Eq.~\eqref{eq:FVD}. These metrics measure visual realism , however as  may ignore physical plausibility or planning relevance \cite{Baraldi2025}.

\begin{equation}
\label{eq:FID}
\mathrm{FID}(\mathcal{X}_r, \mathcal{X}_g)
=
\|\boldsymbol{\mu}_r - \boldsymbol{\mu}_g\|_2^2
+
\mathrm{Tr}
\left(
\boldsymbol{\Sigma}_r
+
\boldsymbol{\Sigma}_g
-
2\left(
\boldsymbol{\Sigma}_r \boldsymbol{\Sigma}_g
\right)^{\frac{1}{2}}
\right)
\end{equation}

\begin{equation}
\label{eq:FVD}
\mathrm{FVD}(\mathcal{V}_r, \mathcal{V}_g)
=
\|\boldsymbol{\mu}_r - \boldsymbol{\mu}_g\|_2^2
+
\mathrm{Tr}
\left(
\boldsymbol{\Sigma}_r
+
\boldsymbol{\Sigma}_g
-
2\left(
\boldsymbol{\Sigma}_r \boldsymbol{\Sigma}_g
\right)^{\frac{1}{2}}
\right)
\end{equation}

Given that:
\begin{itemize}
    \item $\mathcal{X}_r$ and $\mathcal{X}_g$ denote the sets of real and generated image samples, respectively.
    \item $\mathcal{V}_r$ and $\mathcal{V}_g$ denote the sets of real and generated video samples, respectively.
    \item $\boldsymbol{\mu}_r \in \mathbb{R}^d$ and $\boldsymbol{\mu}_g \in \mathbb{R}^d$ represent the empirical mean vectors of deep feature embeddings extracted from real and generated samples.
    \item $\boldsymbol{\Sigma}_r \in \mathbb{R}^{d \times d}$ and $\boldsymbol{\Sigma}_g \in \mathbb{R}^{d \times d}$ denote the empirical covariance matrices of the corresponding feature embeddings.
    \item $\|\cdot\|_2$ denotes the Euclidean ($\ell_2$) norm.
    \item $\mathrm{Tr}(\cdot)$ denotes the trace operator of a square matrix.
    \item $(\boldsymbol{\Sigma}_r \boldsymbol{\Sigma}_g)^{\frac{1}{2}}$ denotes the matrix square root of the product of the two covariance matrices.
    \item $d$ denotes the dimensionality of the feature embedding space.
\end{itemize}

Trajectory metrics based as ADE - Average Distance Error, and FDE - Final Distance Error measure average error but can miss long-tail risk as \cite{Caesar2021,Baraldi2025} suggests. In additional, planning-centric evaluation measures downstream driving performance, but it requires closed-loop testing and can be confounded by simulator limitations. Recent works therefore argue for multi-axis evaluation that includes: (i) predictive accuracy and calibration, (ii) robustness under distribution shift, (iii) usefulness for downstream tasks such as detection, tracking, or planning, and (iv) safety-critical stress tests \cite{Caesar2021,Baraldi2025}.

Datasets and benchmarks shape progress by determining what is measurable. Waymo Open provides scale and diversity for training and evaluating models that must handle real sensor noise and complex urban scenarios \cite{Sun2020}. Cooperative datasets such as V2V4Real add the challenges of multi-agent fusion, localization error, and communication constraints \cite{Xu2023}. Accident-focused benchmarks like DeepAccident stress rare-event anticipation and interaction under occlusion \cite{WangDeepAccident2023}. VRU-focused datasets like OnSiteVRU emphasize mixed traffic and fine-grained interactions \cite{Yan2025}. Together, these resources suggest that “general” driving world models must learn from diverse data sources and must be evaluated across diverse tasks.

The reality gap between simulation and the real world remains a central concern. Simulation allows scalable closed-loop testing, but it may underrepresent rare behaviors or sensor artifacts. Several works therefore emphasize sim-to-real adaptation, hybrid training (real + synthetic), and evaluation protocols that detect overfitting to simulator biases \cite{Xu2023,Li2024}. Framework discussions and surveys highlight the need for standardized reporting and reproducibility across toolchains, since seemingly small implementation choices can dominate conclusions \cite{Chen2025,Jia2025}.

\subsection{Additional Consideration: Surveys, Physics, and Platform Considerations}

Several additional threads in the provided corpus help connect autonomous-driving world models to broader scientific and engineering questions. First, surveys and position pieces emphasize that ``world modeling'' is an umbrella term spanning representations, learning objectives, and downstream uses, and they argue that progress depends on principled choices about what is modeled explicitly versus implicitly as \cite{DawidLeCun2024, LeCun2022}. Recent survey-style works discuss how physical knowledge (dynamics, constraints, and causal structure) can be embedded into learned representations to improve robustness and interpretability \cite{Chen2025,Chu2025}. These discussions are particularly relevant in driving because failures often arise from violations of basic physical plausibility (e.g., impossible motion, inconsistent occlusion) or from spurious correlations in training data - and ones solution suggested by \cite{kong_2015_kinematic, sobal_2022_separating} where integrate primitive and deterministic physical model of the world as microscopic approximation is considerable for learning the physical-affordance aligment as the purpose that world-model that is aim-at \cite{gibson_1979_the}.

Furthermore, works of \cite{Brockman2016,Tassa2018,Audinys2025,Bouzaiene2025} strongly argue that evaluation in embodied domains is inseparable from the environment and platform used for training and testing. Practical benchmarking choices—sensor suites, map availability, traffic participant diversity, and even simulator fidelity—affect what a world model learns and how it generalizes. Platform- and tooling-oriented perspectives highlight that reproducible research requires careful specification of environments, data pipelines, and evaluation procedures. In driving, this connects to the well-known tension between rich simulation for closed-loop testing and real-world data for realism; which does suggest the novel hybrid pipelines that use simulation to explore long-tail scenarios and real data to anchor the model to real sensor statistics, which attempts in mimicing human's capability of acknowledging the physical-affordance \cite{gibson_1979_the}. 

Finally, works as \cite{DawidLeCun2024,LeCun2022, Assran2025} in the World-Modeling showed that, difference modalities can lead to effective representations for downstream and continuous task training, however, such internal representation shall be roburst under enviroment stochasticity and unpredictability, which can be resolved by sequential-based reasoning model with latent-overshooting as \cite{hafner_2019_learning}.

\subsection{Design Implications}

Across these literatures, several design lessons emerge, which does become our result theoretical foundation; First, predictive representation learning that operates in embedding space (e.g., JEPA-style objectives) is increasingly favored over pixel-level reconstruction in driving, because it better matches partial observability and the capability of learning the latent-representation across multi-modal futures \cite{LeCun2022,Zhu2025}.Second, spatially grounded states such as BEV and occupancy are practical and planning-aligned world states, particularly when fused with map priors and explicit modeling of free space \cite{Li2022,Li2025,Bogdoll2025}.Third, multi-agent interaction and V2X communication are becoming core requirements: a driving world model must reconcile distributed observations and predict joint futures under delay and uncertainty \cite{Xu2022,Xu2023,Tu2025}. Finally, safety demands benchmarks that target rare events, accident prediction, shall be balance with efficiency and performance based metrics as reward  is a challenges need to be address in later studies \cite{WangDeepAccident2023,Yan2025,Baraldi2025}.

\section{Proposed World Model and System Architecture}
\label{sec:architecture}

This section shall focus in suggestion in world model contruction, the whole agent interaction flow, and training algorithm that shall be leveraged

\subsection{Notation}
Our problem can be formalizing as the deterministic Markov Decision Process (MDP) with the usage of internal continuous stochastic states and derterministic state, and yeilding the deterministic continuous action in the finite-horizontal enviroment as \cite{sun_2022_supervised, Hafner2023} suggested. The MDP can be defined as the tuples in Eq.~\eqref{eq:MDP}

\begin{equation}
\label{eq:MDP}
\mathcal{M} = (\mathcal{S}, \mathcal{A}, \mathcal{P}, r, \gamma),
\end{equation}

where $\mathcal{S}$ denotes the (possibly continuous) state space, 
$\mathcal{A}$ denotes the (possibly continuous) action space, 
$\mathcal{P}(s_{t+1} \mid s_t, a_t)$ represents the state transition dynamics which including the stochastic states $s_t$ capturing the enviroment randomness pattern within the transition, and determistic states $h_t$ yeild from the recurrent-planner for historical context,
$r : \mathcal{S} \times \mathcal{A} \rightarrow \mathbb{R}$ is the reward function,
and $\gamma \in [0,1)$ is the discount factor - and for the setting we are adopt dirrectly the setting of DreamerV3 by setting  $\gamma$ = 0.997.

The action space we attempts including the ego's vehicle acceleration, and steering angle (in radian) as the response for making the interaction to the simmulation.

\subsection{Overall System Architecture}
\begin{figure}[t]
    \centering
    \includegraphics[width=0.85\linewidth]{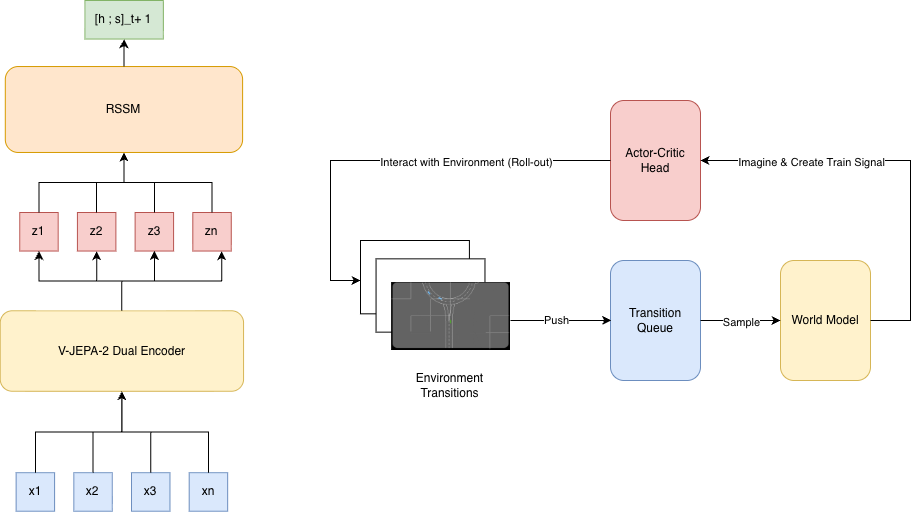}
    \caption{The proposed HanoiWorld World Model (left) include an visual encoder based on V-JEPA-2 checkpoint proposed by \cite{assran_2025_vjepa} and the RSSM-backbone suggested by \cite{hafner_2019_learning} for making long-term planning on the next possible transition of enviroment - the green block. The overal system arrchitectural of the enviroment on the (right) which been design aiming for both effective rolling-out while model training from the enviroment by creating such feedback loop on the agent interaction with data-scarcity}
    \label{fig:architecture_and_world_model}
\end{figure}

The proposed overall system archtiecture include modules as enviroment interface, which build based on the HighwayEnv package provided by \cite{faramafoundation_2025_gymnasium} for agent interaction, we additionally including the Transtion Queues with limited in size as the experience collector for the World Model can be sampling randomly certain episode from the queue that have with satisfied sequence length $T$, the World Model shall provide the imaginary embedding as the train-signal toward the Actor-Critic controler head for yielding the action for reacting toward the enviroment and create an close interacting cycle until the agent reach the terminatin states, the architecture of the whole HanoiWorld is presented on the right side of Figure~\ref{fig:architecture_and_world_model}.
    
The transition queue shall storing the list of episode (eg. reference of valid episode) that have the length of transition satisfy certain threadshold - $T$, and the sequence shall contain the Bird-Eye-View RGB images and the metadata on the agent's interaction as the agent's action, ego's actions, and the description on the reward been yield across the simulation for those sequence.

\subsection{HanoiWorld - The World Model}

\subsubsection*{V-JEPA 2 Based Encoder}

The main innovation of the HanoiWorld is the inclusion of the strong pretrained image encoder based on self-supervised learning manner proposed by \cite{assran_2025_vjepa}. Specifically, the V-JEPA 2 aim at learning the essential knowledge on enviroment interaction as the motion, object movement , without focusing on the stochasitic and noisy pixel detail as reconstruction attemps as DreamerV3-based encoders \cite{Hafner2023, hu_2023_gaia1}. The V-JEPA 2 encoder shall be trained to predict the masked version of the corresponding input follow self-supervising manner for more 1 million hours of video, and be stablized using Exponential Moving Average (EMA) in the Student-Teacher scheme for preventing the embedding collaspe while maintaining essential feature are strucutrally preserved \cite{caron_2021_emerging}.

\begin{figure}[t]
    \centering
    \includegraphics[width=0.85\linewidth]{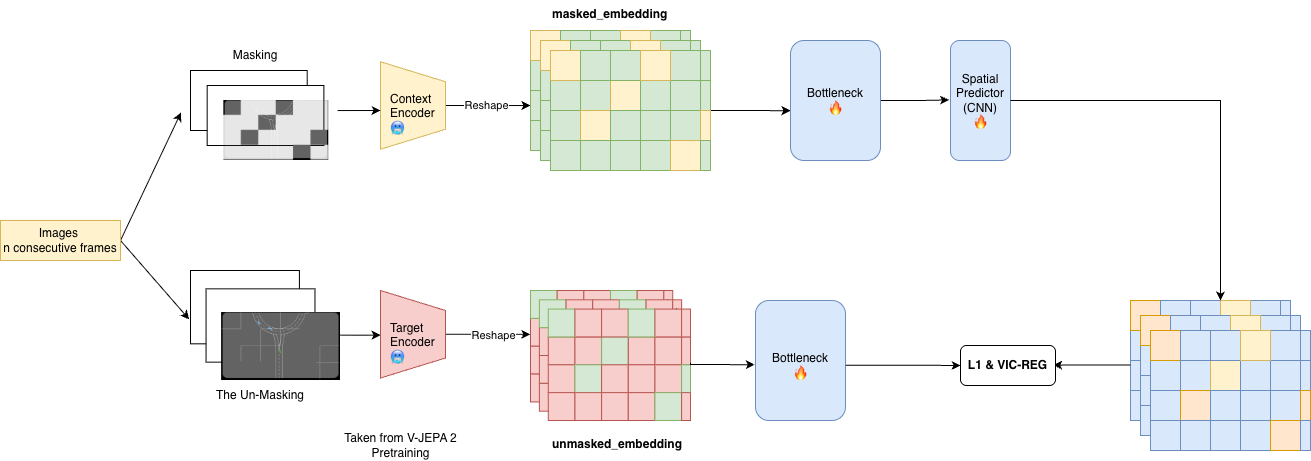}
    \caption{Overview of the proposed HanoiWorld encoder architecture.
    A pretrained and frozen V-JEPA~2 encoder~\cite{assran_2025_vjepa} is used as a high-quality representation backbone to improve training efficiency and embedding robustness under limited-data settings.
    A downstream bottleneck Multi-Layer Perceptron (MLP) is trained to project the high-dimensional representations into a compact and task-compatible latent space of size $1024 \times 128$.
    In parallel, the student encoder branch incorporates an additional 2D convolutional neural network (CNN) module to predict spatial representations.
    Both branches are jointly optimized using an $\ell_1$ alignment loss as Eq.~\eqref{eq:jepa_loss} and the VICReg regularization objective~\cite{bardes_2022_vicreg}.}
    \label{fig:encoder_architecture}
\end{figure}

\paragraph{EMA Teacher Encoder.}
Let $\theta$ denote the parameters of the student encoder $E_\theta$
and $\bar{\theta}$ denote the parameters of the teacher encoder.
The teacher parameters are updated using an exponential moving average:
\begin{equation}
\bar{\theta}
\;\leftarrow\;
\tau\,\bar{\theta}
+
(1 - \tau)\,\theta,
\label{eq:ema_update}
\end{equation}
where $\tau \in [0,1)$ is the momentum coefficient.
Gradients are stopped through the teacher encoder to prevent representation collapse, and in our experiment, we set $\tau$ = 0.996.

\paragraph{Masked Representation Prediction.}
Given a video $y$ and a masked view $x$ obtained by removing a subset of spatio-temporal patches,
the encoder $E_\theta$ extracts representations from the visible tokens.
A predictor $P_\phi$ is trained to predict the representations of the masked tokens.

The main training objective is defined as the L1-Loss as the Eq.~\eqref{eq:jepa_loss}:

\begin{equation}
\mathcal{L}_{\text{align}}
=
\left\|
P_\phi\!\left(\Delta_y, E_\theta(x)\right)
-
\operatorname{sg}\!\left(E_{\bar{\theta}}(y)\right)
\right\|_1,
\label{eq:jepa_loss}
\end{equation}

where $\Delta_y$ denotes learnable mask tokens indicating the locations of the masked patches,
and $\operatorname{sg}(\cdot)$ denotes the stop-gradient operator.
The loss is applied only to the masked patches, and we leverage the patch-mask with random-masking as \cite{mo_2024_connecting} suggest for the finetuning process.

The leverage of the dual-branch self-supervised training with masking pattern does occur in work of \cite{zhu_2025_selfsupervised}, in which suggest that by let the encoder have to force to guess the masked patch from the sampel - which does creating challenge for preventing embedding collaspe while occupancy semenatic can be learn and predicted without the need of full geometrical structure on the enviroment.

\subsubsection*{RSSM-Based Reasoning Model}

The RSSM (Recurrent State-Space Model) our design is follow the codebase provided with the DreamerV3 architecture by \cite{hafner_2019_learning, Hafner2023}, the purpose of RSSM considered as the internal-memory module that storing the deterministic memory for long-term historical semantics, and the stochastic latent values as the uncertainty prediction based on the enviroment encoded-signals for capturing the world's evolution, and agent's expected prior conditioned by the world's state transformation. The generative process on yeilding the World Dynamics can be formulated as the follow Eq.~\eqref{eq:rssm_generative}

\begin{equation}
\label{eq:rssm_generative}
\begin{aligned}
h_t &= f_\theta(h_{t-1}, z_{t-1}, a_{t-1}), \\
z_t &\sim p_\theta(s_t \mid h_t), \\
z_t &\sim p_\theta(z_t \mid h_t, z_t), \\
r_t &\sim p_\theta(r_t \mid h_t, z_t).
\end{aligned}
\end{equation}

Given that:
\begin{itemize}
    \item $f_\theta$ denotes the recurrent dynamics function parameterized by $\theta$,
    implemented as a recurrent neural network.
    \item $h_t$ represents the deterministic latent state at time step $t$.
    \item $a_t$ denotes the action executed by the agent at time step $t$.
    \item $z_t$ corresponds to the observation at time step $t$, which in our setting
    is the V-JEPA-2 encoded embedding for the corresponding time step.
    \item $r_t$ denotes the predicted reward yielded by the agent at time step $t$.
    \item All conditional distributions $p_\theta(\cdot)$ are parameterized by
    multi-layer perceptron (MLP) decoder networks.
\end{itemize}

Additionally, as the HanoiWorld migrating the RSSM-codebased from the DreamerV3, the model does additional including a continue-predictor for guessing whether the episode shall be continue given the encoder embedding $o_t$, and historical latent images $z_t$ \cite{Hafner2023} - showcased with Eq.~\eqref{eq:continue_predictor}, given the continuation signal is a Bernouli random variable for simplified assumption.

\begin{equation}
\label{eq:continue_predictor}
c_t \sim p_\theta(c_t \mid h_t, z_t),
\end{equation}
where $c_t \in \{0,1\}$ indicates whether the episode continues at time step $t$.

The continue-predictor training as the logistic-regression model using the binary-cross-entropy as the original DreamerV3 suggestion with the Eq.~\eqref{eq:continuation_loss}
\begin{equation}
\label{eq:continuation_loss}
\mathcal{L}_{\text{cont}}
=
-
\left[
c_t \log \hat{c}_t
+
(1 - c_t)\log(1 - \hat{c}_t)
\right].
\end{equation}

\begin{itemize}
    \item $c_t \in \{0,1\}$ denotes the ground-truth binary continuation label at time step $t$.
    \item $\hat{c}_t \in (0,1)$ denotes the predicted continuation probability produced by the model.
\end{itemize}

\subsection{The Actor-Critic Training Head}

The HanoiWorld Actor--Critic framework is directly derived from the reward- and value-based learning paradigm of DreamerV3. It shares core conceptual foundations with the classical theory proposed by \cite{sutton_1991_dyna}, which emphasizes a separation between an \emph{Actor} network (policy network) that reacts by selecting actions and a \emph{Critic} network that evaluates the agent's performance through a value function in order to support more effective planning. However, unlike the classical Actor--Critic formulation in \cite{sutton_1991_dyna}, which operates on actual environment states, the DreamerV3 Actor--Critic operates entirely in a learned latent environment. Specifically, both the policy network $\pi_\theta$ and the value network $v_\psi$ are defined over the latent state $\ell_t$ produced by the RSSM, as formalized in Eq.~\eqref{eq:actor_critic_latent}.

\begin{equation}
\label{eq:actor_critic_latent}
\ell_t = (h_t, z_t), \qquad
a_t \sim \pi_\theta(a_t \mid \ell_t), \qquad
v_\psi(R_t \mid \ell_t).
\end{equation}

Given that:
\begin{equation}
R_t = \sum_{k=0}^{\infty} \gamma^k r_{t+k}.
\end{equation}

\begin{itemize}
    \item $\ell_t = (h_t, z_t)$ denotes the latent state at time step $t$,
    consisting of the deterministic and stochastic components of the RSSM and V-JEPA-2 encoder respectively.
    \item $R_t$ denotes the return at time step $t$, defined as the discounted
    sum of future rewards over an episode,
    with discount factor $\gamma = 0.997$.
\end{itemize}

\subsection{Training Procerdure an Objective }

HanoiWorld training objective including the Spatial Predictor and Predictor training in the Encoder module, and training the RSSM-dynamics based model, with the Actor-Critic training follow. Given that HanoiWorld training tactics only mimicking the training procedure of DreamerV3 on the Dynamics, and Actor-Critic network, not the reconstructed-based training for encoder as \cite{Hafner2023}.

\subsubsection*{Encoder Training}

For the Bottleneck and Spatial Predictor training, beside using the L1-loss as Eq.~\eqref{eq:jepa_loss}, we consider the usage of Variance Correlation and Covariance Regularization suggested by \cite{bardes_2022_vicreg} - check Eq.~\eqref{eq:var} and Eq.~\eqref{eq:covar}

\begin{equation}
\label{eq:var}
\mathcal{L}_{\text{var}}
=
\frac{1}{D}
\sum_{d=1}^{D}
\max\!\left(0,\;
1 - \sqrt{\operatorname{Var}\!\left(\tilde{z}_{:,d}\right) + \varepsilon}
\right),
\end{equation}

\begin{equation}
\label{eq:covar}
\mathcal{L}_{\text{cov}}
=
\frac{1}{D}
\sum_{i \neq j}
\left(
\operatorname{Cov}(\tilde{z})_{ij}
\right)^2,
\end{equation}

\begin{equation}
\label{eq:covar_mat}
\operatorname{Cov}(\tilde{z})
=
\frac{1}{BN - 1}
\left(
\tilde{\mathbf{z}} - \bar{\mathbf{z}}
\right)^{\!\top}
\left(
\tilde{\mathbf{z}} - \bar{\mathbf{z}}
\right),
\end{equation}

Given that:

\begin{itemize}
    \item $D$ denotes the embedding dimensionality (D=128), and
    $\tilde{\mathbf{z}} \in \mathbb{R}^{(BN)\times D}$ represents the flattened
    embeddings obtained by reshaping the batch and token dimensions.
    \item $\tilde{z}_{:,d}$ refers to the $d$-th embedding dimension across
    all $BN$ samples, and $\varepsilon$ is a small constant added for numerical
    stability.
    \item The variance regularization in Eq.~\eqref{eq:var} enforces a minimum
    standard deviation for each embedding dimension, preventing representational
    collapse.
    \item $\epsilon$ standfor the small constant for numerical stability as $\epsilon$ = 1e-4
    
    \item The covariance matrix $\operatorname{Cov}(\tilde{z})$ in
    Eq.~\eqref{eq:covar_mat} is computed after centering the embeddings by their
    mean $\bar{\mathbf{z}} = \frac{1}{BN}\sum_{i=1}^{BN}\tilde{\mathbf{z}}_i$.
    \item The covariance regularization in Eq.~\eqref{eq:covar} penalizes the
    squared off-diagonal entries of the covariance matrix, encouraging
    decorrelation between different embedding dimensions.
\end{itemize}

The Bottleneck are train that follow the weighted sum loss function as follow \cite{bardes_2022_vicreg, assran_2025_vjepa, zhu_2025_selfsupervised}, and with the weights are setup toward  ($\alpha$, $\beta$, $\gamma$) to (1.0, 1.0, 0.1) for stability and prevent collaspe, and lossely controlling the covariance between teacher's bottleneck and student's predictor.

\begin{equation}
\mathcal{L}_{\text{encoder}}
=
\alpha\,\mathcal{L}_{\text{align}}
+
\beta\,\mathcal{L}_{\text{var}}
+
\gamma\,\mathcal{L}_{\text{cov}}.
\end{equation}

We does finetune our encoder on the 2D Bird-Eye-View dataset which have been pre-processed from the Nuscene dataset proposed by \cite{Caesar2021} by rendering from the LiDAR-cloud based with additional metadata on the obstacle and enviroment movement toward the RGB based images that V-JEPA 2 encoder can working with.

\subsubsection*{RSSM-dynamic model training}

For the RSSM module training, we leverage the usage of predictor loss which is the inverse summation on the log probability of the probability in guessing the actual observation on the enviroment's transition, the reward function, and continue flag as Eq~\eqref{eq:L_pred}; the dynamicity loss by minimalizing the KL-diveregnce between actual posterior distribution encoder yeild out from actual observation stably, with prior that world model imagine from prior-latent memory as Eq~\eqref{eq:L_dyn}; and the representation loss for anchoring the encoder without drifting away from latent world model expectation in Eq~\eqref{eq:L_rep}. This loss-setup follows the work of \cite{Hafner2023} suggested and using the identical weight-set as DreamerV3 been configured before.

\begin{align}
\mathcal{L}_{\text{pred}}(\phi)
&=
- \log p_\phi(x_t \mid z_t, h_t)
- \log p_\phi(r_t \mid z_t, h_t)
- \log p_\phi(c_t \mid z_t, h_t),
\label{eq:L_pred}
\\[6pt]
\mathcal{L}_{\text{dyn}}(\phi)
&=
\max\!\left(
1,\;
\mathrm{KL}\!\left(
\operatorname{sg}\!\left[q_\phi(z_t \mid h_t, x_t)\right]
\;\big\|\;
p_\phi(z_t \mid h_t)
\right)
\right),
\label{eq:L_dyn}
\\[6pt]
\mathcal{L}_{\text{rep}}(\phi)
&=
\max\!\left(
1,\;
\mathrm{KL}\!\left(
q_\phi(z_t \mid h_t, x_t)
\;\big\|\;
\operatorname{sg}\!\left[p_\phi(z_t \mid h_t)\right]
\right)
\right).
\label{eq:L_rep}
\end{align}

\subsubsection*{REINFORCE-based Actor-Critic learning}

The algorithm HanoiWorld Actor-Critic component are based on the Reinforce algorithm suggest by \cite{williams_1992_simple}, which suggest the agent (specific the policy network of the Actor) shall optimize and favor the action that yeilded best return. However,the algorithm we used follow the implementation of DreamerV3 from \cite{Hafner2023}, which suggest both actor and critic with imagined roll out from latent RSSMs, with the critic layer yeild the prediction on the possible cummulative reward based on the latent feature on the future, while the actor's policy network are optimized based on the advantage signal , and the entropy regularization for trigger agent exploration on potential action with imaginative near-future - and the entropy are being scale by a fixed constant.The actor critics algorithm training is explicit describe on the Algorithm~\ref{alg:actor_critic}.

\begin{algorithm}[htbp]
\caption{Actor--Critic Training via Imagined Rollouts}
\label{alg:actor_critic}
\KwIn{
World model parameters $\phi$;\\
Actor parameters $\theta$;\\
Critic parameters $\psi$;\\
Imagined horizon $H$;\\
Discount factor $\gamma$;\\
$\lambda$-return parameter $\lambda$;\\
Entropy coefficient $\beta$ = 3e-4
}

Sample posterior latent state $(h_0, z_0)$ from real experience\;
Initialize imagined trajectory buffers\;

\For{$t = 0$ \KwTo $H-1$}{
    Compute feature $f_t \leftarrow f_\phi(h_t, z_t)$\;
    Sample action $a_t \sim \pi_\theta(\cdot \mid f_t)$\;
    Predict next latent state
    $(h_{t+1}, z_{t+1}) \sim p_\phi(\cdot \mid h_t, z_t, a_t)$\;
}

\end{algorithm}

\begin{algorithm}[htbp]
\caption*{Algorithm~\ref{alg:actor_critic} (continued)}

\For{$t = 0$ \KwTo $H-1$}{
    Predict reward $\hat r_{t+1} \leftarrow r_\phi(h_{t+1}, z_{t+1})$\;
    Predict continuation $\hat c_{t+1} \leftarrow c_\phi(h_{t+1}, z_{t+1})$\;
}

Compute value predictions $V_t \leftarrow V_\psi(f_t)$ for all $t \in \{0,\dots,H\}$\;

\BlankLine
\textbf{Compute $\lambda$-return targets}\;
Set effective discount $\gamma_t \leftarrow \gamma \cdot \hat c_t$\;
Set bootstrap target $G_H^\lambda \leftarrow V_H$\;


\For{$t = H-1, H-2, \dots, 0$}{
    $G_t^\lambda \leftarrow
    \hat r_{t+1}
    +
    \gamma_{t+1}
    \bigl(
        (1-\lambda)\, V_{t+1}
        +
        \lambda\, G_{t+1}^\lambda
    \bigr)$\;
}

\BlankLine
\textbf{Critic update}\;
Minimize negative log-likelihood of $\lambda$-returns:\;
$\mathcal{L}_{\text{value}}(\psi)
\leftarrow
-\,\mathbb{E}_t\!\left[
\log p_\psi(G_t^\lambda \mid f_t)
\right]$\;

\BlankLine
\textbf{Actor update}\;
Compute advantage (baseline):
$A_t \leftarrow \operatorname{sg}(G_t^\lambda - V_t)$\;

Minimize actor loss:\;
$\mathcal{L}_{\text{actor}}(\theta)
\leftarrow
-\,\mathbb{E}_t\!\left[
\log \pi_\theta(a_t \mid f_t)\, A_t
+
\beta\, \mathcal{H}\!\left(\pi_\theta(\cdot \mid f_t)\right)
\right]$\;

Update critic parameters $\psi$\;
Update actor parameters $\theta$\;

\end{algorithm}

\section{Experimental Result}
\label{sec:experiment}
This section of the paper shall focusing on the experiment that been used for evaluating the peformance of the HanoiWorld in comparison with difference baseline approachs

\subsection{Experiment Description}

For evaluating the HanoiWorld performance in comparison toward across difference alternatives, experiments shall be conducted within the enviroment from the HighwayEnv package provided by \cite{faramafoundation_2025_gymnasium}, these include \textit{Highway, Roundabout, Merge}, which shall be discussed in detail in subsection ~\ref{sssec:casestudy}

Comparative World Model design that we choosen in designing are including the DreamerV3 proposed by \cite{Hafner2023}, the VQ-VAE + ConvLSTM based suggested by \cite{robine_2020_smaller}, and the HanoiWorld prosposed agents. The baseline model described as below:
\begin{itemize}
    \item The VQ-VAE encoder based with ConvLSTM planner suggested by \cite{robine_2020_smaller} is a small and discrete latent world model which aimming for efficient inference and planning on the latent space, as the inheritence on the idea of \cite{ha_2018_world}. The VQ-VAE based solution is design on 100 thoudsand episode on the game of Atari by using the Proximal Policy Optimization (PPO) algorithm prosposed by \cite{schulman_2017_proximal} for latent space planning on discrete imagined rollouts.
    \item The DreamerV3 by \cite{Hafner2023} currently is reached State-Off-The-Art level performance with the capability of long-term hozirontal planning by the RSSMs networks, which does inspired the design of HanoiWorld by attemtps on learning the roburst, and generalizing model for enviroment agnostic, which use to predict the plausible next latent state, reward, and continuation, the model is leveraged for yielding the training signal to multi-layer-perceptron layers on reward, and values prediction based on imagined rollout, with Actor follow the REINFORCE gradient update, or hybridizing update learning signal from RSSM for robust and stable learning
\end{itemize}

For the evaluation metric, the selected metric 2 metric - the average reward (e.g score), and the collision rate as the following:
\begin{itemize}

\item \textit{The average reward/score} is the metric that \cite{ha_2018_world} used for evaluating the effeciency of the agents on planing efficency - which is the average of the reward signal across steps over episodes - which showcased with the Eq.~\eqref{eq:avg_reward}
\item \textit{The Collision Rate} is the metric suggested by \cite{gao_2024_cardreamer} which can be estimated by the proportion of episode that ego vehicles occur the collision, which expressed with Eq.~\eqref{eq:collision_rate}
\end{itemize}
\begin{equation}
\label{eq:avg_reward}
\bar{R}
=
\frac{1}{N}
\sum_{i=1}^{N}
\sum_{t=1}^{T_i}
r_{i,t},
\end{equation}

\noindent\textbf{Given that:}
\begin{itemize}
    \item $\bar{R}$ denotes the average driving reward over all evaluation episodes.
    \item $N$ is the total number of evaluation episodes.
    \item $i \in \{1, \dots, N\}$ indexes the episode.
    \item $t$ indexes the time steps within an episode.
    \item $r_{i,t}$ is the instantaneous reward received at time step $t$ in episode $i$.
\end{itemize}

\noindent\textbf{Interpretation:}
Higher values of $\bar{R}$ indicate better driving behavior, meaning smoother control (less wobbling)
and safer trajectories (fewer crashes).
\begin{equation}
\label{eq:collision_rate}
\text{Collision Rate}
=
\frac{\#\,\text{collision episodes}}
{\text{total episodes that being evaluated}}
\end{equation}

\subsection{Case Studies}
\label{sssec:casestudy}

The experiment had been conducted on the interactive enviroment for examinining the model's adaptability and generalizabilities across difference driving situations as the difference road-topology, lane merging scenarios, and navigation in the roundabout.

The information and tabular desription of each scenarios are provided within the table~\ref{table:env_config}.

\begin{sidewaystable}[!p]
\centering
\small
\renewcommand{\arraystretch}{1.25}
\begin{tabular}{lccc}
\hline
\textbf{Aspect}
& \textbf{Highway}
& \textbf{Merge}
& \textbf{Roundabout} \\
\hline

Task
& \texttt{highway\_highway}
& \texttt{highway\_merge}
& \texttt{highway\_roundabout} \\

Observation
& RGB image ($64 \times 64$)
& RGB image ($64 \times 64$)
& RGB image ($64 \times 64$) \\

Action space
& Continuous
& Discrete
& Discrete \\

Episode time limit
& 200
& 800
& 800 \\

Traffic setting
& Dense straight highway
& On-ramp merging traffic
& Circular multi-agent traffic \\

Vehicle count
& 50
& Environment-controlled
& Environment-controlled \\

Vehicle density
& 1.5
& Implicit (spawn-based)
& Implicit (spawn-based) \\

Speed target (m/s)
& 23--27
& 20--28
& 8--15 \\

Collision penalty
& $-5.0$
& $-1.0$
& $-1.0$ \\

Safe-distance shaping
& Strong penalty (tailgating)
& Reward + penalty
& Reward + penalty (stronger) \\

Lane-change shaping
& Smart lane change rewarded
& Lane changes penalized
& Slight penalty \\

Progress shaping
& Moderate
& Strong
& Moderate \\

Heading alignment reward
& Yes
& Yes
& Yes (strong) \\

Survival reward
& -- 
& Low
& Moderate \\

Success reward
& 1.0
& 0.8
& 1.0 \\

Reward shaping weight
& 0.85
& 0.8
& 0.8 \\

Imagination gradient
& REINFORCE
& REINFORCE
& REINFORCE \\

Primary behavioral objective
& Stable high-speed driving
& Safe and anticipatory merging
& Social negotiation and yielding \\

\hline
\end{tabular}
\caption{Comparison of Highway-based autonomous driving environments and their core configuration and reward-shaping characteristics.}
\label{table:env_config}
\end{sidewaystable}

\subsubsection*{Task description on each enviroment}
\begin{figure}[h]
    \centering
    \includegraphics[width=0.85\linewidth]{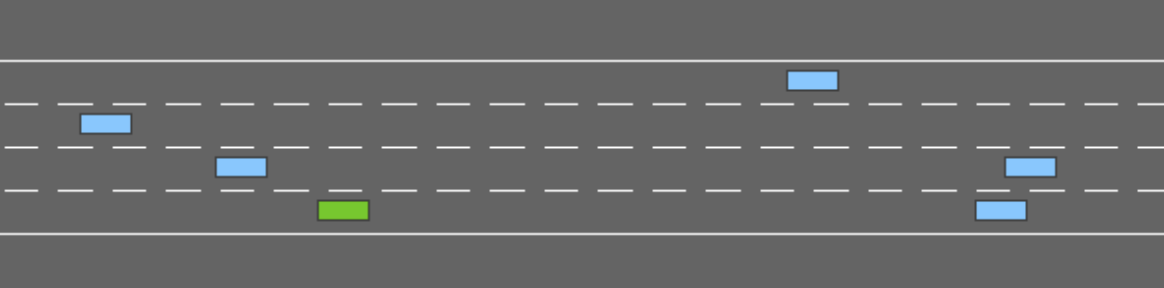}
    \caption{The highway-v0 Enviroment - a snapshot taken from \cite{faramafoundation_2025_gymnasium} }
    \label{fig:highway}
\end{figure}

The highway-env (e.g highway-v0) - Figure~\ref{fig:highway} is designed to showcase the model's capability of controlling the vehicle's movement on multi-lane scenarioes where ego vehicle shall interact and make lane changing decision in corresponse to difference agent's (surrounding vehicles can accelerate or suddently de-accelerate, and change-lanes). The enviroment attempts to test the model's capability on maintain stablizing, and safe-driving speed, and the flexibility by making lane switch as the correspondence toward surrounding agents \cite{faramafoundation_2025_gymnasium}.

\begin{figure}[h]
    \centering
    \includegraphics[width=0.85\linewidth]{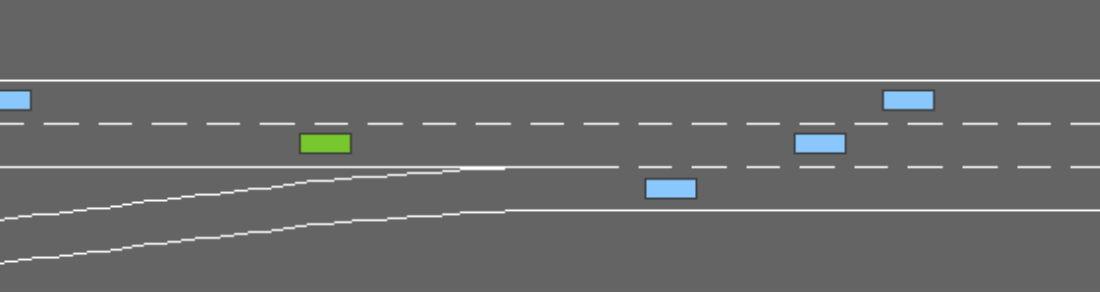}
    \caption{The roundabout-v0 Enviroment - a snapshot taken from \cite{faramafoundation_2025_gymnasium} }
    \label{fig:merge}
\end{figure}

The merge-env (e.g merge-v0) - Figure~\ref{fig:merge}, within the simulation, the agent shall have to make the attempts on switching from 1 road-branch toward difference for addressing the lane-merging challenge, which is challenging dues to the speed controlling, and detail planning of the agent in preventing the collision during the merging process \cite{faramafoundation_2025_gymnasium}. The goal of the scenario is the model attempts understand and predicted the quick physical transition on difference agents for making appropriate merging-decision.

\begin{figure}[h]
    \centering
    \includegraphics[width=0.85\linewidth]{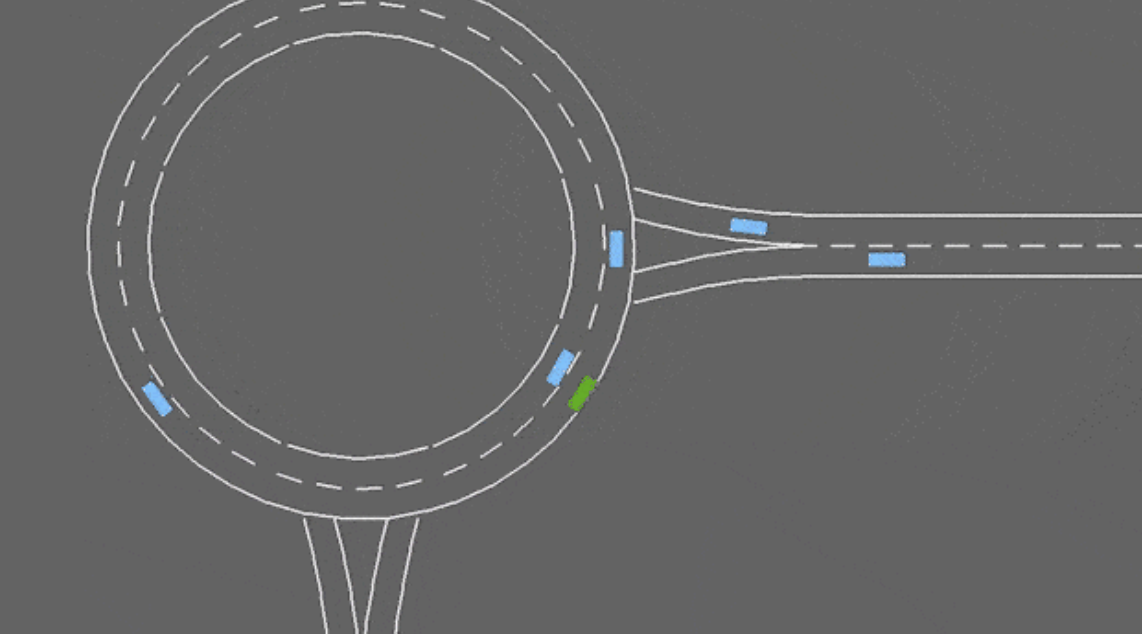}
    \caption{The roundabout-v0 Enviroment - a snapshot taken from \cite{faramafoundation_2025_gymnasium} }
    \label{fig:roundabout}
\end{figure}

The roundabout (roundabout-v0) as Figure~\ref{fig:roundabout} is the complex and realistic version of the merge-v0 where the agents have more than 1 suggestion in navigation, and the density of the navigation across each dirrection is stochastic as the prior usecases. The purpose of this situation is considered as the roburstness checking on the performance with more closer toward realistic driving  enviroment, where the ego's shall make the planning on when to navigate, stop to entering the roundabout without trggering an accident \cite{faramafoundation_2025_gymnasium}.

\subsection{Experimental Procedure}
For the baseline - both the DreamerV3, and VQ-VAE based model shall be trained within 100 thousand steps including the rolling-out for training based on the sample; while HanoiWorld dues to the larger model-structure but the RSSM-planning module are still effective in the inference - it shall be trained under 5000 steps with the prior-pretrained V-JEPA 2 with the bottleneck and spatial embedding predictor. All model world-model (except for the HanoiWorld encoder) shall be trained primitively within the Highway-env simmulation.

On the evaluation phase of the experiment, difference model shall be evaluating and infered within 100 vary length episode on 3 difference enviroments, and returning the average reward/score and the overall collision rate within 100 episodes, and the result shall be record both the mean-values, and the standard deviation of the measurement across episodes.

\subsection{Evaluation}
The experiment result shall be showcased with the table~\ref{tab:collision_rate} for collision rate, and table~\ref{tab:avg_reward} for average reward scoring

\begin{table}[!h]
\centering
\caption{Collision rate over 100 evaluation episodes (mean $\pm$ std).}
\label{tab:collision_rate}
\begin{tabular}{lccc}
\hline
\textbf{ENV} & \textbf{highway-v0} & \textbf{merge-v0} & \textbf{roundabout-v0} \\
\hline
DreamerV3   & $0.550 \pm 0.497$ & $0.030 \pm 0.171$ & $0.500 \pm 0.500$ \\
VQ-VAE      & $1.000 \pm 0.000$ & $0.290 \pm 0.454$ & $0.570 \pm 0.495$ \\
HanoiWorld & $0.200 \pm 0.400$ & $0.970 \pm 0.170$ & $0.340 \pm 0.473$ \\
\hline
\end{tabular}
\end{table}

\begin{table}[!h]
\centering
\caption{Average episode reward over 100 evaluation episodes (mean $\pm$ std).}
\label{tab:avg_reward}
\begin{tabular}{lccc}
\hline
\textbf{ENV} & \textbf{highway-v0} & \textbf{merge-v0} & \textbf{roundabout-v0} \\
\hline
DreamerV3   & $51.065 \pm 52.700$ & $41.973 \pm 2.896$ & $9.423 \pm 6.171$ \\
VQ-VAE      & $3.121 \pm 12.047$  & $30.114 \pm 11.664$ & $3.826 \pm 4.643$ \\
HanoiWorld & $13.163 \pm 23.277$ & $13.480 \pm 5.703$ & $9.818 \pm 6.252$ \\
\hline
\end{tabular}
\end{table}

From the perspective of the collision rate, HanoiWorld demonstrates superior performance compared to the corresponding baselines on the Highway and Roundabout environments, with average collision rates of $0.200$ and $0.340$, respectively. This outperforms DreamerV3, which exhibits mid-tier collision performance, while VQ-VAE performs worst among the evaluated methods. However, HanoiWorld reveals a notable weakness in the lane-merging scenario, where it exhibits the highest expected collision rate - while DreamerV3 show it's stronger capability of navigation in this situation with the rate of $0.030$. From the model's stability - which is expressed through the collision rate stability, our HanoiWorld does show statistical stability over domain-wise with lowest standard deviation, however these does show the signal that HanoiWorld can under poor navigation performance in merge-v0 scenario where the collision rate near to 1.

Extend toward the average episode reward, the HanoiWorld only showcased it's performance competively with SOTA baselines of DreamerV3 under the roundabout-v0 enviroments with certain raise on the average reward $9.818$ with $9.423$ respectively, whereas,  the DreamerV3 still show the effectively planning efficency with the highest planning performance, with the highway-v0 show significant stable planning then merge-v0 scenarios, while the HanoiWorld performance under these 2 task is in the middle on the highway case; and poor planning in the merge-v0 simulation even with smaller parameter model as VQ-VAE baselines.

\subsection{Results Discussion}

As the result been showcased, the HanoiWorld can effectively planning on the both simple (highway-v0) and complex enviroment (roundabout-v0) with the competive performance, while it's showcased such poor generalization with mordering challenging as merging - where the frequencies of collision is substantially high, while the performance on reward yielding as ones could observe is consistence across enviroment, which proof the HanoiWorld can generalizing the planning tatic across difference enviroment. However, on the merge-v0 we assume the model have the signal of reward hacking by favoring the action of collision in yeilding consistence reward across episode.

In additionally, as we observe, in the case where model need to showcased the flexibility in road navigation as roundabout-v0 in lane selection, the model currently favor the safe-lane selection, which does show that our entropy based regularization for making decision is underperformed and need to be carefully studies with abalation.

Even though HanoiWorld showcases competitive and worth-considering performance in the comparison with state-of-the-art baselines as DreamerV3, there are certain aspect that this studies can be addressed within the nearby future, as the more global-contextual introduction on the enviroment for gating on the affordance relationship between the enviroment and the agent's perspective by using text-conditioned language encoder as \cite{hu_2023_gaia1}, or occupancies encoder on the whole enviroment graph as the \cite{Zhao2024, vasudevan_2024_planning} for generalzing behavior; additionally the studies does not inspect the imaginary rolling out as the work for \cite{ha_2018_world} -- which can consider for later studies on checking the performance of the ego's with difference imaginary temperature for hardness controlling as the migration for reward-regularization within the latent domain.
\section{Conclusion}
\label{sec:conclusion}

This study showcased HanoiWorld, a JEPA-based worldmodel for simmulation generation in training autonomous vehicle controller in reinforcement learning, as the result have proved HanoiWorld is capable in making effective planning strategy within the latent domain as the SOTA baseline, and lighter competitor, make the attemptss seem compelling than the pixel-reconstruction based approach dues to computational efficency. Additionally, our model does show that in some certain scenarios HanoiWorld reach lower collision rate, which can hypothesize the model do learn the concept of safetyness both across enviroment, however the planning mechanism of the model are under-performing in comparing with SOTA baseline as DreamerV3.

Remarkably, the current study does not attempt in the integration of more global contextual condition using multi-modality inputs as the language, global graph; in addition with the latent-probing for actual evaluating -- understanding the planing mechanism of the WorldModel under difference enviromental hardness; suggesting the potential of further dirrection that later study can be inspired and continues using HanoiWorld's result.

\newpage
\bibliographystyle{unsrt}
\bibliography{cite}

@article{Assran2025,
  title     = {Self-Supervised Learning from Images with a Joint-Embedding Predictive Architecture},
  author    = {Assran, Mahmoud and Bardes, Adrien and Ponce, Jean and LeCun, Yann},
  journal   = {arXiv preprint arXiv.2301.08243},
  year      = {2025}
}

@misc{assran_2025_vjepa,
  author = {Assran, Mido and Bardes, Adrien and Fan, David and Garrido, Quentin and Howes, Russell and Komeili, Mojtaba and Muckley, Matthew and Rizvi, Ammar and Roberts, Claire and Sinha, Koustuv and Zholus, Artem and Arnaud, Sergio and Gejji, Abha and Martin, Ada and Hogan, Francois Robert and Dugas, Daniel and Bojanowski, Piotr and Khalidov, Vasil and Labatut, Patrick and Massa, Francisco and Szafraniec, Marc and Krishnakumar, Kapil and Li, Yong and Ma, Xiaodong and Chandar, Sarath and Meier, Franziska and LeCun, Yann and Rabbat, Michael and Ballas, Nicolas},
  title = {V-JEPA 2: Self-Supervised Video Models Enable Understanding, Prediction and Planning},
  year = {2025},
  organization = {arXiv.org}
}

@article{Brockman2016,
  title={OpenAI Gym},
  author={Brockman, Greg and Cheung, Vicki and Pettersson, Ludwig and Schneider, Jonas and Schulman, John and Tang, Jie and Zaremba, Wojciech},
  journal={arXiv preprint arXiv:1606.01540},
  year={2016}
}

@misc{jia_2023_adriveri,
  author = {Jia, Fan and Mao, Weixin and Liu, Yingfei and Zhao, Yucheng and Wen, Yuqing and Zhang, Chi and Zhang, Xiangyu and Wang, Tiancai},
  month = {10},
  title = {ADriver-I: A General World Model for Autonomous Driving},
  urldate = {2025-11-02},
  year = {2023},
  organization = {arXiv.org}
}

@misc{zhang_2024_bevworld,
  author = {Zhang, Yumeng and Gong, Shi and Xiong, Kaixin and Ye, Xiaoqing and Li, Xiaofan and Tan, Xiao and Wang, Fan and Huang, Jizhou and Wu, Hua and Wang, Haifeng},
  month = {06},
  title = {BEVWorld: A Multimodal World Simulator for Autonomous Driving via Scene-Level BEV Latents},
  urldate = {2025-11-02},
  year = {2024},
  organization = {arXiv.org}
}

@article{dulacarnold_2019_challenges,
  author = {Dulac-Arnold, Gabriel and Mankowitz, Daniel and Hester, Todd},
  month = {04},
  title = {Challenges of Real-World Reinforcement Learning},
  year = {2019},
  journal = {arXiv:1904.12901 [cs, stat]}
}

@article{ha_2018_world,
  author = {Ha, David and Schmidhuber, Jürgen},
  pages = {e10},
  title = {World Models},
  doi = {10.5281/zenodo.1207631},
  volume = {1},
  year = {2018},
  journal = {World Models}
}

@misc{sun_2022_supervised,
  author = {Sun, Hao and Xu, Ziping and Fang, Meng and Zhou, Bolei},
  title = {Supervised Q-Learning can be a Strong Baseline for Continuous Control},
  urldate = {2025-11-12},
  year = {2022},
  organization = {Openreview.net}
}

@article{Tassa2018,
  title={DeepMind Control Suite},
  author={Tassa, Yuval and others},
  journal={arXiv preprint arXiv:1801.00690},
  year={2018}
}

@article{Hafner2023,
  title={Mastering Diverse Domains through World Models},
  author={Hafner, Danijar and others},
  journal={arXiv preprint arXiv:2301.04104},
  year={2023}
}

@misc{mo_2024_connecting,
  author = {Mo, Shentong and Tong, Shengbang},
  month = {10},
  title = {Connecting Joint-Embedding Predictive Architecture with Contrastive Self-supervised Learning},
  urldate = {2025-11-06},
  year = {2024}
}

@article{sutton_1991_dyna,
  author = {Sutton, Richard},
  month = {07},
  pages = {160-163},
  publisher = {Association for Computing Machinery},
  title = {Dyna, an integrated architecture for learning, planning, and reacting},
  doi = {10.1145/122344.122377},
  volume = {2},
  year = {1991},
  journal = {SIGART newsletter}
}

@misc{zhu_2025_selfsupervised,
  author = {Zhu, Haoran and Dong, Zhenyuan and Topollai, Kristi and Sha, Beiyao and Choromanska, Anna},
  title = {Self-Supervised Representation Learning with Joint Embedding Predictive Architecture for Automotive LiDAR Object Detection},
  urldate = {2025-11-02},
  year = {2025},
  organization = {arXiv.org}
}

@article{caron_2021_emerging,
  author = {Caron, Mathilde and Touvron, Hugo and Misra, Ishan and Jégou, Hervé and Mairal, Julien and Bojanowski, Piotr and Joulin, Armand},
  month = {05},
  title = {Emerging Properties in Self-Supervised Vision Transformers},
  year = {2021},
  journal = {arXiv:2104.14294 [cs]}
}

@misc{hu_2023_gaia1,
  author = {Hu, Anthony and Russell, Lloyd and Yeo, Hudson and Murez, Zak and Fedoseev, George and Kendall, Alex and Shotton, Jamie and Corrado, Gianluca},
  month = {09},
  title = {GAIA-1: A Generative World Model for Autonomous Driving},
  year = {2023},
  organization = {arXiv.org}
}

@misc{faramafoundation_2025_gymnasium,
  author = {Farama Foundation},
  title = {Gymnasium Documentation},
  year = {2025},
  organization = {gymnasium.farama.org}
}

@article{LeCun2022,
  title={A Path Towards Autonomous Machine Intelligence},
  author={LeCun, Yann},
  journal={arXiv preprint arXiv:2203.03620},
  year={2022}
}

@article{Caron2021,
  title={Emerging Properties in Self-Supervised Vision Transformers},
  author={Caron, Mathilde and others},
  journal={ICCV},
  year={2021}
}

@article{ZhengVedaldi2023,
  title={Online Clustered Codebook},
  author={Zheng, Chuanxia and Vedaldi, Andrea},
  journal={arXiv preprint arXiv:2307.15139},
  year={2023}
}

@article{Zhu2025,
  title={Self-Supervised Representation Learning with Joint Embedding Predictive Architecture for Automotive LiDAR Object Detection},
  author={Zhu, Haoran and others},
  journal={arXiv preprint arXiv:2501.04969},
  year={2025}
}

@article{Sun2020,
  title={Scalability in Perception for Autonomous Driving: Waymo Open Dataset},
  author={Sun, Pei and others},
  journal={CVPR},
  year={2020}
}

@misc{gao_2024_cardreamer,
  author = {Gao, Dechen and Cai, Shuangyu and Zhou, Hanchu and Wang, Hang and Soltani, Iman and Zhang, Junshan},
  title = {CarDreamer: Open-Source Learning Platform for World Model based Autonomous Driving},
  urldate = {2025-10-24},
  year = {2024},
  organization = {arXiv.org}
}

@article{Caesar2021,
  title={nuScenes: A Multimodal Dataset for Autonomous Driving},
  author={Caesar, Holger and others},
  journal={CVPR},
  year={2021}
}

@article{Xu2021,
  title={OPV2V: An Open Benchmark Dataset and Fusion Pipeline for Perception with Vehicle-to-Vehicle Communication},
  author={Xu, Runsheng and others},
  journal={ICRA},
  year={2021}
}

@article{Xu2022,
  title={V2X-ViT: Vehicle-to-Everything Cooperative Perception with Vision Transformer},
  author={Xu, Runsheng and others},
  journal={ECCV},
  year={2022}
}

@article{Xu2023,
  title={V2V4Real: A Real-World Large-Scale Dataset for Vehicle-to-Vehicle Cooperative Perception},
  author={Xu, Runsheng and others},
  journal={CVPR},
  year={2023}
}

@article{WangDeepAccident2023,
  title={DeepAccident: A Motion and Accident Prediction Benchmark for V2X Autonomous Driving},
  author={Wang, Shaoshuai and others},
  journal={NeurIPS},
  year={2023}
}

@article{Zhao2024,
  title={DriveDreamer-2: LLM-Enhanced World Models for Diverse Driving Video Generation},
  author={Zhao, Guosheng and others},
  journal={arXiv preprint arXiv:2403.06845},
  year={2024}
}

@misc{vasudevan_2024_planning,
  author = {Vasudevan, Arun Balajee and Peri, Neehar and Schneider, Jeff and Ramanan, Deva},
  title = {Planning with Adaptive World Models for Autonomous Driving},
  urldate = {2025-10-24},
  year = {2024},
  organization = {arXiv.org}
}

@article{Gao2024,
  title={CarDreamer: Learning World Models for Autonomous Driving},
  author={Gao, Rui and others},
  journal={arXiv preprint arXiv:2406.10101},
  year={2024}
}

@article{Fu2024,
  title={Driving Diffusion Models},
  author={Fu, Yiheng and others},
  journal={arXiv preprint arXiv:2405.02345},
  year={2024}
}

@article{Li2022,
  title={BEVFormer: Learning Bird's-Eye-View Representation from Multi-Camera Images},
  author={Li, Yinhao and others},
  journal={ECCV},
  year={2022}
}

@article{Li2024,
  title={Occupancy Prediction for Autonomous Driving},
  author={Li, Zhi and others},
  journal={CVPR},
  year={2024}
}

@article{Li2025,
  title={LiDAR-Centric World Models for Autonomous Driving},
  author={Li, Wei and others},
  journal={arXiv preprint},
  year={2025}
}

@article{Bogdoll2025,
  title={Occupancy-Based World Modeling for Safety-Critical Autonomous Driving},
  author={Bogdoll, Julian and others},
  journal={arXiv preprint},
  year={2025}
}

@misc{Baraldi2025,
 author = {Baraldi, Lorenzo and Zeng, Zifan and Zhang, Chongzhe and Nayak, Aradhana and Zhu, Hongbo and Liu, Feng and Zhang, Qunli and Wang, Peng and Liu, Shiming and Hu, Zheng and Cangelosi, Angelo and Baraldi, Lorenzo},
  title = {The Safety Challenge of World Models for Embodied AI Agents: A Review},
  urldate = {2025-12-31},
  year = {2025},
  organization = {arXiv.org}
}

@article{Yan2025,
  title={OnSiteVRU: A High-Resolution Trajectory Dataset for Vulnerable Road Users},
  author={Yan, Zhangcun and others},
  journal={arXiv preprint arXiv:2503.23365},
  year={2025}
}

@article{Minh2022,
  title={Masked Autoencoders for Point Clouds},
  author={Minh, Nguyen and others},
  journal={ECCV},
  year={2022}
}

@article{Min2024,
  title={Occupancy-MAE: Self-Supervised Learning for Autonomous Driving},
  author={Min, Chen and others},
  journal={CVPR},
  year={2024}
}

@article{hafner_2019_learning,
  author = {Hafner, Danijar and Lillicrap, Timothy and Fischer, Ian and Villegas, Ruben and Ha, David and Lee, Honglak and Davidson, James},
  month = {06},
  title = {Learning Latent Dynamics for Planning from Pixels},
  year = {2019},
  journal = {arXiv:1811.04551 [cs, stat]}
}

@article{Tu2025,
  title={Delay-Aware Cooperative Perception for Autonomous Driving},
  author={Tu, Xiaoyu and others},
  journal={arXiv preprint},
  year={2025}
}

@article{Tu2025b,
  title={Robust Cooperative World Models under Communication Constraints},
  author={Tu, Xiaoyu and others},
  journal={arXiv preprint},
  year={2025}
}

@article{Vasudevan2024,
  title={Transformer-Based Planning for Autonomous Driving},
  author={Vasudevan, Siva and others},
  journal={CVPR},
  year={2024}
}

@article{Wu2025,
  title={Scaling World Models for Autonomous Driving},
  author={Wu, Yifan and others},
  journal={arXiv preprint},
  year={2025}
}

@article{Xie2025,
  title={Unified World Modeling and Planning for Autonomous Driving},
  author={Xie, Yiming and others},
  journal={arXiv preprint},
  year={2025}
}

@article{Long2025,
  title={Interaction-Aware World Models for Dense Traffic},
  author={Long, Han and others},
  journal={arXiv preprint},
  year={2025}
}

@article{Feng2025,
  title={Multi-Agent Intent Modeling for Autonomous Driving},
  author={Feng, Qian and others},
  journal={arXiv preprint},
  year={2025}
}

@article{DawidLeCun2024,
  title={Energy-Based World Models},
  author={Dawid, Alexandre and LeCun, Yann},
  journal={arXiv preprint},
  year={2024}
}

@article{Audinys2025,
  title={Benchmarking World Models for Autonomous Systems},
  author={Audinys, Lukas and others},
  journal={arXiv preprint},
  year={2025}
}

@article{Bouzaiene2025,
  title={Simulation Platforms for Autonomous Driving Research},
  author={Bouzaiene, Khaled},
  journal={arXiv preprint},
  year={2025}
}

@article{Guan2025,
  title={Large-Scale Training of Driving World Models},
  author={Guan, Yifan and others},
  journal={arXiv preprint},
  year={2025}
}

@article{Hansen2023,
  title={Generalization in Model-Based Reinforcement Learning},
  author={Hansen, Nicklas and others},
  journal={NeurIPS},
  year={2023}
}

@article{Jia2025,
  title={Survey of World Models in Autonomous Driving},
  author={Jia, Ming and others},
  journal={arXiv preprint},
  year={2025}
}

@misc{morimura_2024_policy,
  author = {Morimura, Tetsuro and Ota, Kazuhiro and Abe, Kenshi and Zhang, Peinan},
  title = {Policy Gradient Algorithms with Monte Carlo Tree Learning for Non-Markov Decision Processes},
  urldate = {2025-11-18},
  year = {2024}
}

@misc{robine_2020_smaller,
  author = {Robine, Jan and Uelwer, Tobias and Harmeling, Stefan},
  title = {Smaller World Models for Reinforcement Learning},
  urldate = {2025-11-23},
  year = {2020},
  organization = {arXiv.org}
}

@misc{xiao_2019_learning,
  author = {Xiao, Chenjun and Wu, Yifan and Ma, Chen and Schuurmans, Dale and Müller, Martin},
  title = {Learning to Combat Compounding-Error in Model-Based Reinforcement Learning},
  urldate = {2025-12-29},
  year = {2019},
  organization = {arXiv.org}
}

@misc{momchil_2025_treeirl,
  author = {Momchil , Tomov S and Lee, Sang Uk and Hendrago, Hansford and Huh, Jinwook and Han, Teawon and Howington, Forbes and Silva, da and Bernasconi, Gianmarco and Heim, Marc and Findler, Samuel and Ji, Xiaonan and Boule, Alexander and Napoli, Michael and Chen, Kuo and Miller, Jesse and Floor, Boaz and Hu, Yunqing},
  title = {TreeIRL: Safe Urban Driving with Tree Search and Inverse Reinforcement Learning},
  urldate = {2025-11-12},
  year = {2025},
  organization = {arXiv.org}
}

@misc{huang_2024_learning,
  author = {Huang, Zhiyu and Tang, Chen and Lv, Chen and Tomizuka, Masayoshi and Zhan, Wei},
  title = {Learning Online Belief Prediction for Efficient POMDP Planning in Autonomous Driving},
  urldate = {2025-11-18},
  year = {2024},
  organization = {arXiv.org}
}

@misc{han_2024_a,
  author = {Han, Ye and Zhang, Lijun and Meng, Dejian and Zhang, Zhuang and Hu, Xingyu and Weng, Songyu},
  title = {A Value Based Parallel Update MCTS Method for Multi-Agent Cooperative Decision Making of Connected and Automated Vehicles},
  urldate = {2025-11-12},
  year = {2024},
  organization = {arXiv.org}
}

@article{hoel_2019_combining,
  author = {Hoel, Carl-Johan and Driggs-Campbell, Katherine and Wolff, Krister and Laine, Leo and Kochenderfer, Mykel},
  pages = {1-1},
  title = {Combining Planning and Deep Reinforcement Learning in Tactical Decision Making for Autonomous Driving},
  doi = {10.1109/tiv.2019.2955905},
  urldate = {2020-02-04},
  year = {2019},
  journal = {IEEE Transactions on Intelligent Vehicles}
}

@misc{hessel_2017_rainbow,
  author = {Hessel, Matteo and Modayil, Joseph and Hasselt, van and Schaul, Tom and Ostrovski, Georg and Dabney, Will and Horgan, Dan and Piot, Bilal and Azar, Mohammad and Silver, David},
  title = {Rainbow: Combining Improvements in Deep Reinforcement Learning},
  year = {2017},
  organization = {arXiv.org}
}

@article{Chen2025,
  title={Physical World Modeling for Autonomous Agents},
  author={Chen, Rui and others},
  journal={arXiv preprint},
  year={2025}
}

@misc{kong_2015_kinematic,
  author = {Kong, Jason and Pfeiffer, Mark and Schildbach, Georg and Borrelli, Francesco},
  month = {06},
  pages = {1094–1099},
  title = {Kinematic and dynamic vehicle models for autonomous driving control design},
  doi = {10.1109/IVS.2015.7225830},
  urldate = {2022-12-10},
  year = {2015},
  organization = {IEEE Xplore}
}

@article{bardes_2022_vicreg,
  author = {Bardes, Adrien and Ponce, Jean and LeCun, Yann},
  month = {01},
  title = {VICReg: Variance-Invariance-Covariance Regularization for Self-Supervised Learning},
  year = {2022},
  journal = {arXiv:2105.04906 [cs]}
}

@misc{gibson_1979_the,
  author = {Gibson, James J.},
  title = {THE ECOLOGICAL APPROACH TO VISUAL PERCEPTION},
  year = {1979}
}

@misc{schulman_2017_proximal,
  author = {Schulman, John and Wolski, Filip and Dhariwal, Prafulla and Radford, Alec and Klimov, Oleg},
  month = {08},
  title = {Proximal Policy Optimization Algorithms},
  year = {2017},
  organization = {arXiv.org}
}

@article{williams_1992_simple,
  author = {Williams, Ronald J.},
  month = {05},
  pages = {229-256},
  title = {Simple statistical gradient-following algorithms for connectionist reinforcement learning},
  doi = {10.1007/bf00992696},
  volume = {8},
  year = {1992},
  journal = {Machine Learning}
}

@article{hacohen_2022_autonomous,
  author = {Hacohen, Shlomi and Medina, Oded and Shoval, Shraga},
  month = {05},
  pages = {21241-21258},
  title = {Autonomous Driving: A Survey of Technological Gaps Using Google Scholar and Web of Science Trend Analysis},
  doi = {10.1109/tits.2022.3172442},
  volume = {23},
  year = {2022},
  journal = {IEEE Transactions on Intelligent Transportation Systems}
}

@misc{sobal_2022_separating,
  author = {Sobal, Vlad and Canziani, Alfredo and Carion, Nicolas and Cho, Kyunghyun and LeCun, Yann},
  title = {Separating the World and Ego Models for Self-Driving},
  urldate = {2025-10-30},
  year = {2022},
  organization = {arXiv.org}
}

@article{Chu2025,
  title={World Models: A Survey},
  author={Chu, Yifan and others},
  journal={arXiv preprint},
  year={2025}
}

@article{WangDriveDreamer2023,
  title     = {DriveDreamer: Towards Realistic World Models for Autonomous Driving},
  author    = {Wang, Xiaofeng and Zhao, Guosheng and Zhu, Zheng and Chen, Xinze and Huang, Guan and Bao, Xiaoyi and Wang, Xingang},
  journal   = {arXiv preprint arXiv:2310.05008},
  year      = {2023}
}

\newpage
\clearpage
\appendix
\section{Additional Results}
\label{sec:appendix}


\begin{figure}[h]
    \centering
    \includegraphics[width=0.85\linewidth]{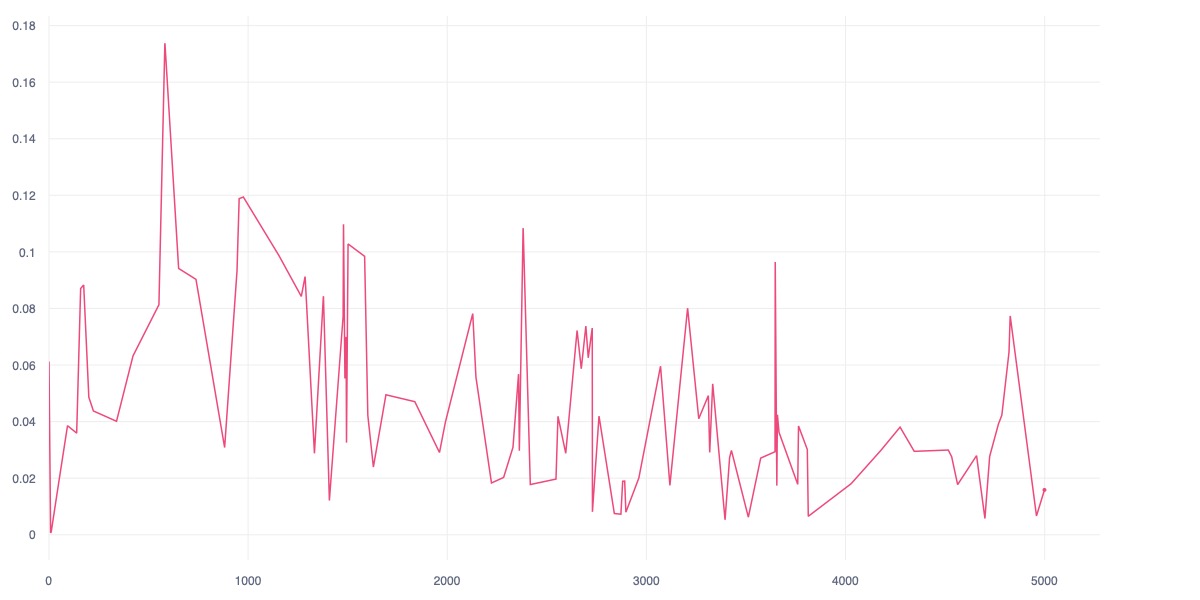}
    \caption{Merge environment: continuation loss of the RSSM as a function of training steps.}
    \label{fig:merge_cont_loss}
\end{figure}

\begin{figure}[h]
    \centering
    \includegraphics[width=0.85\linewidth]{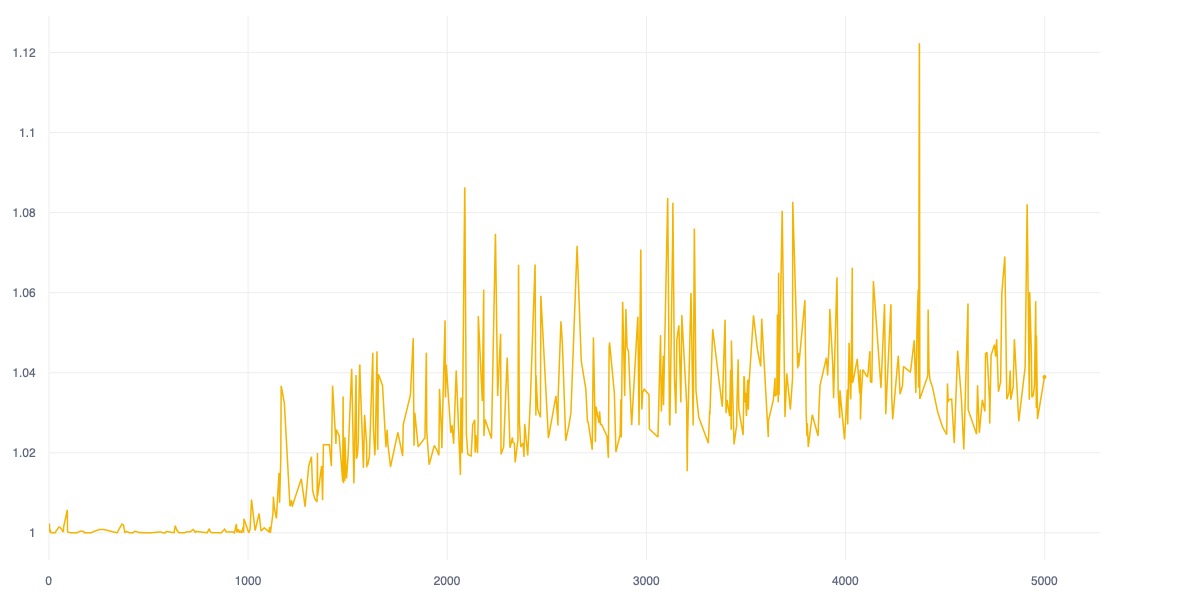}
    \caption{Merge environment: dynamics loss of the RSSM as a function of training steps.}
    \label{fig:merge_dyn_loss}
\end{figure}

\begin{figure}[p]
    \centering
    \includegraphics[width=0.85\linewidth]{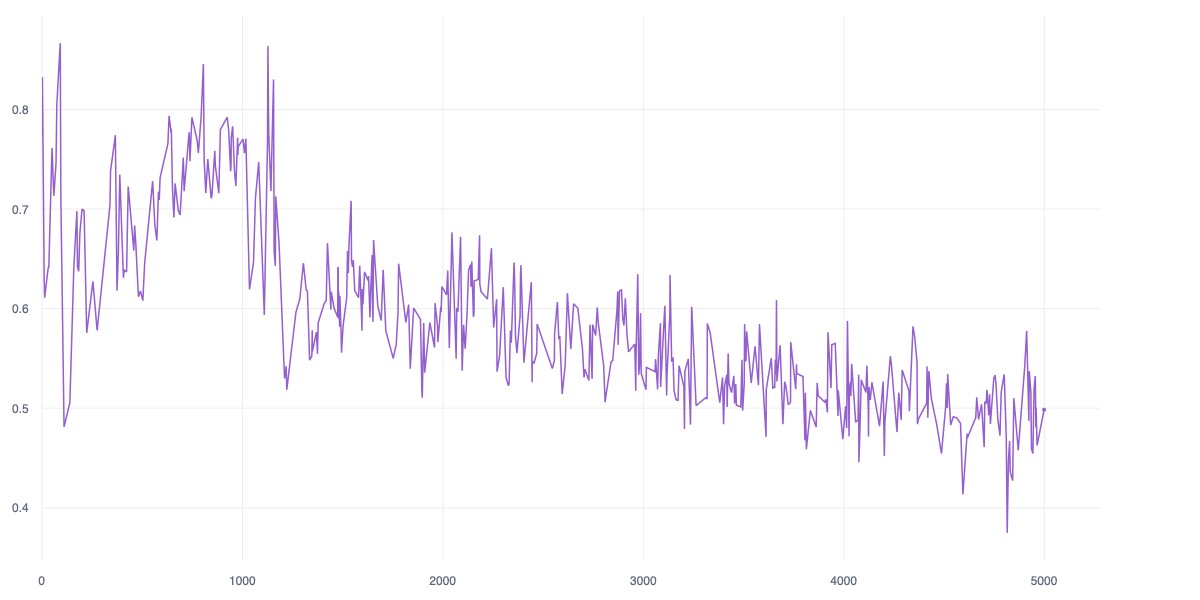}
    \caption{Merge environment: KL regularization loss of the RSSM over training steps.}
    \label{fig:merge_kl_loss}
\end{figure}

\begin{figure}[p]
    \centering
    \includegraphics[width=0.85\linewidth]{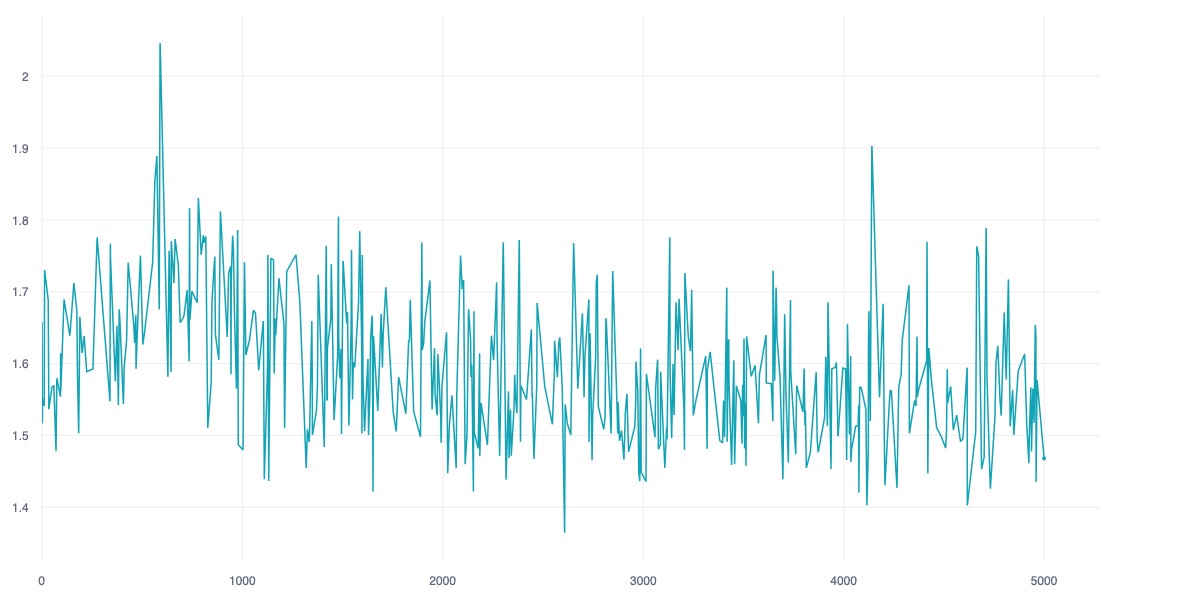}
    \caption{Merge environment: overall HanoiWorld RSSM model loss over training steps.}
    \label{fig:merge_model_loss}
\end{figure}

\begin{figure}[p]
    \centering
    \includegraphics[width=0.85\linewidth]{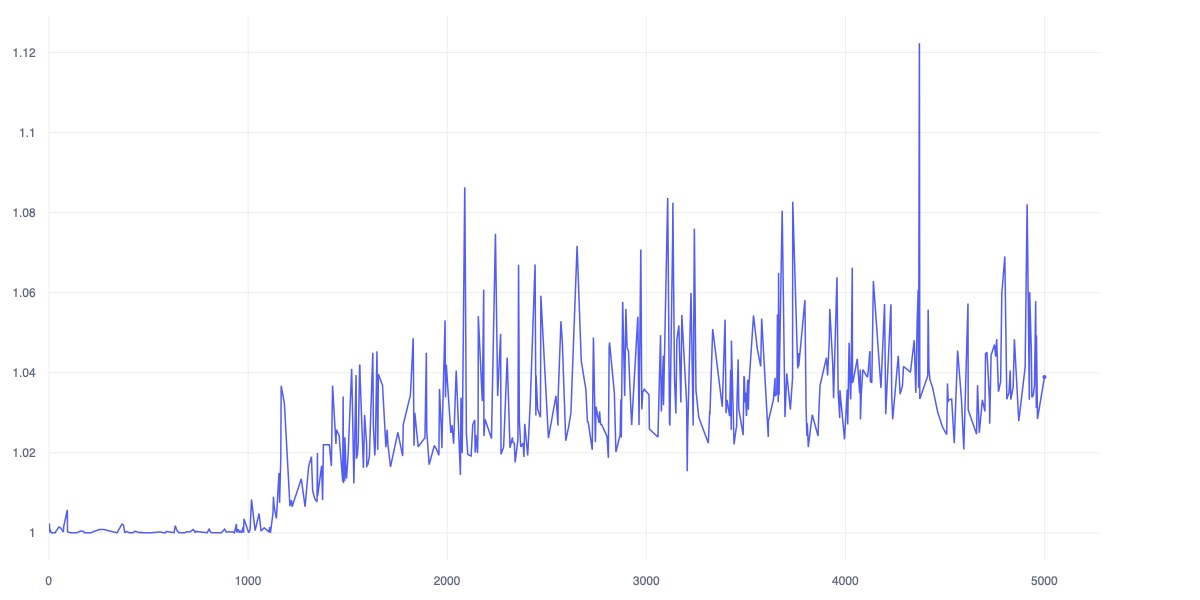}
    \caption{Merge environment: representation loss of the RSSM over training steps.}
    \label{fig:merge_rep_loss}
\end{figure}

\begin{figure}[p]
    \centering
    \includegraphics[width=0.85\linewidth]{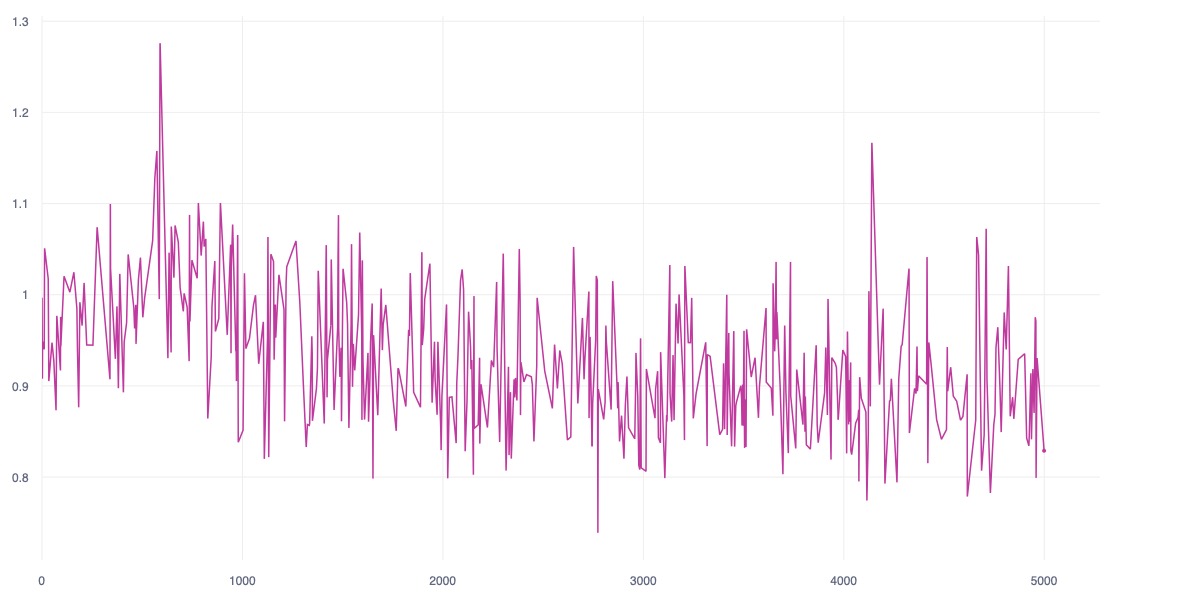}
    \caption{Merge environment: reward prediction loss of the RSSM over training steps.}
    \label{fig:merge_reward_loss}
\end{figure}


\begin{figure}[p]
    \centering
    \includegraphics[width=0.85\linewidth]{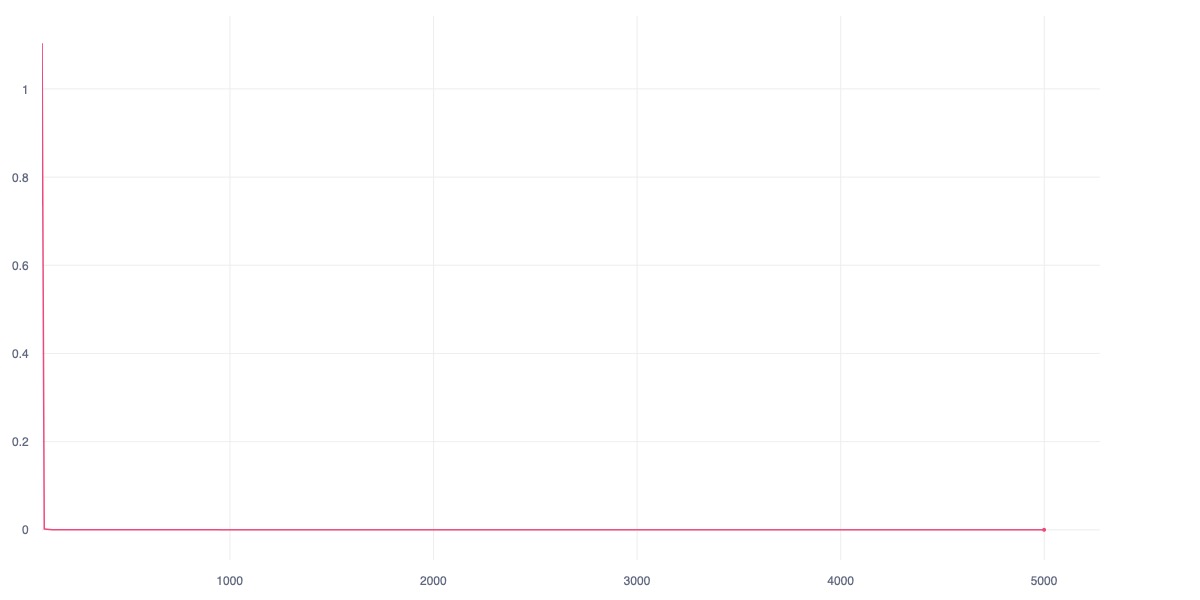}
    \caption{Highway environment: continuation loss of the RSSM as a function of training steps.}
    \label{fig:highway_cont_loss}
\end{figure}

\begin{figure}[p]
    \centering
    \includegraphics[width=0.85\linewidth]{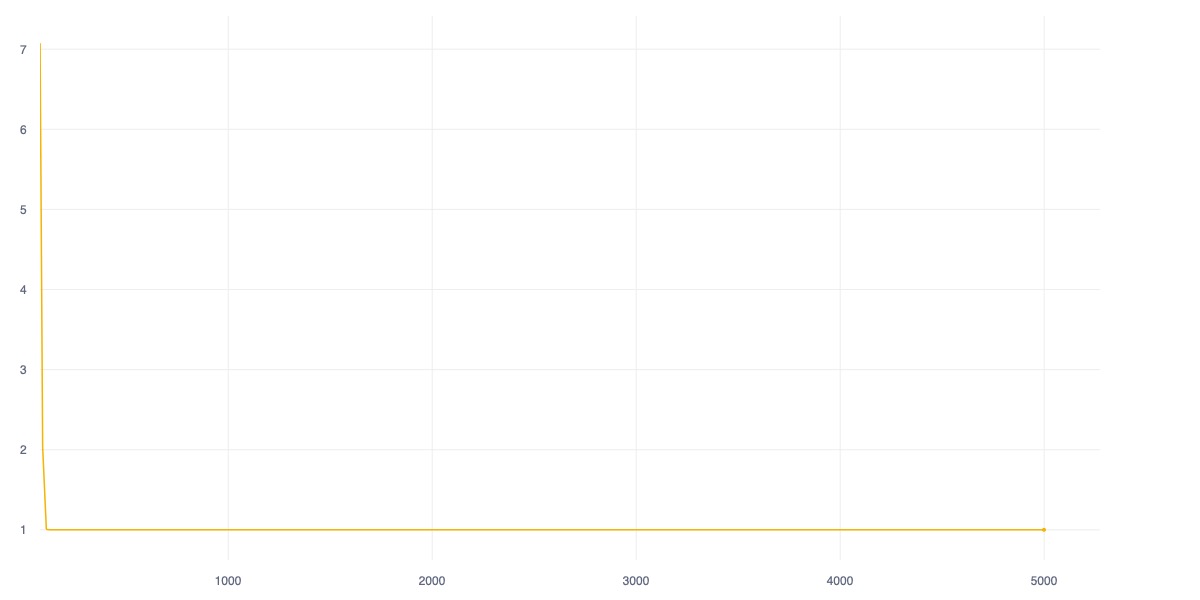}
    \caption{Highway environment: dynamics loss of the RSSM as a function of training steps.}
    \label{fig:highway_dyn_loss}
\end{figure}

\begin{figure}[p]
    \centering
    \includegraphics[width=0.85\linewidth]{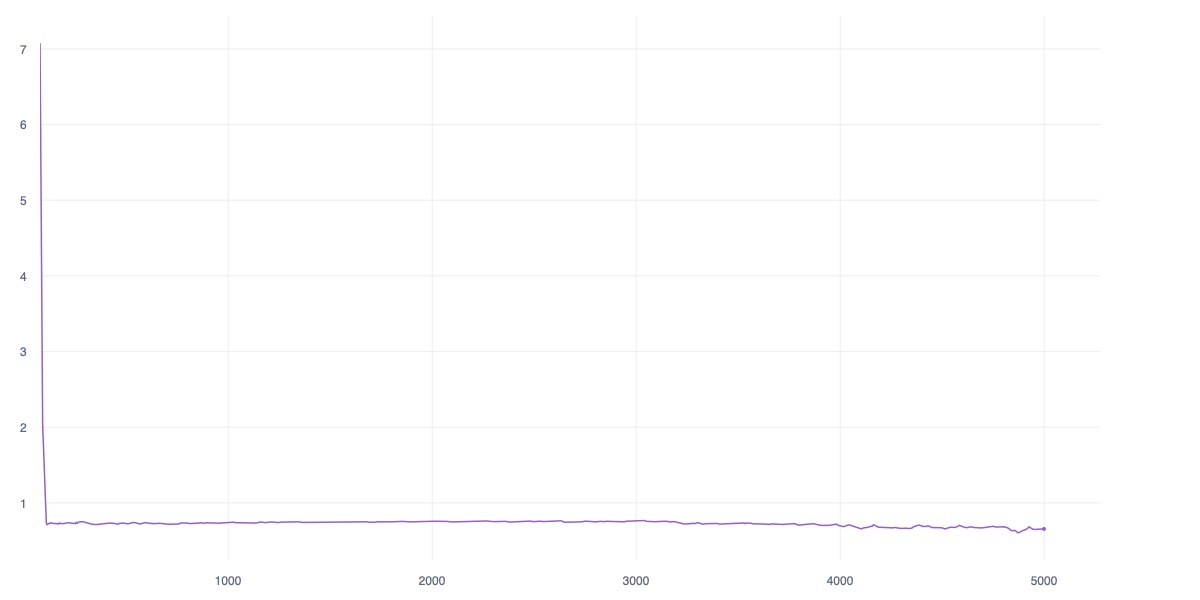}
    \caption{Highway environment: KL regularization loss of the RSSM over training steps.}
    \label{fig:highway_kl_loss}
\end{figure}

\begin{figure}[p]
    \centering
    \includegraphics[width=0.85\linewidth]{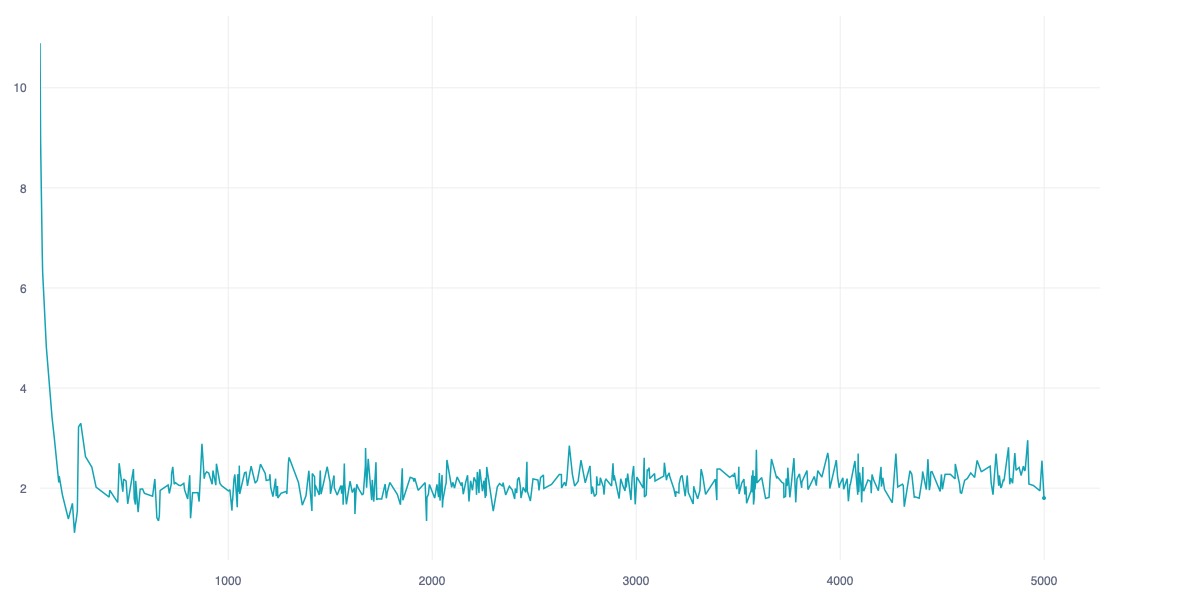}
    \caption{Highway environment: overall RSSM model loss over training steps.}
    \label{fig:highway_model_loss}
\end{figure}

\begin{figure}[p]
    \centering
    \includegraphics[width=0.85\linewidth]{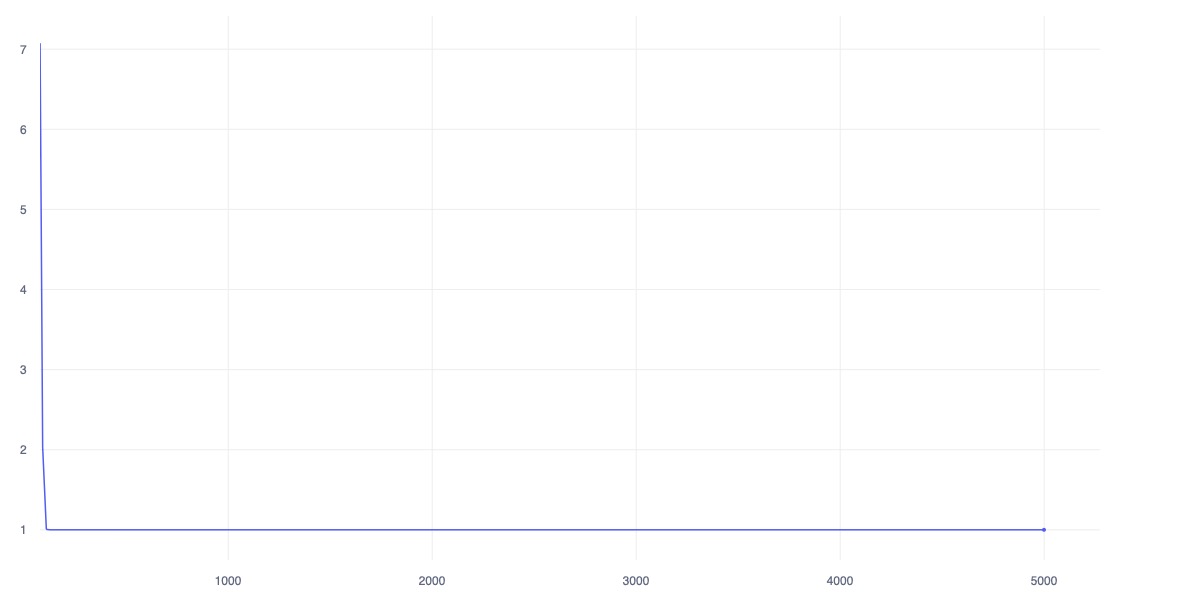}
    \caption{Highway environment: representation loss of the RSSM over training steps.}
    \label{fig:highway_rep_loss}
\end{figure}

\begin{figure}[p]
    \centering
    \includegraphics[width=0.85\linewidth]{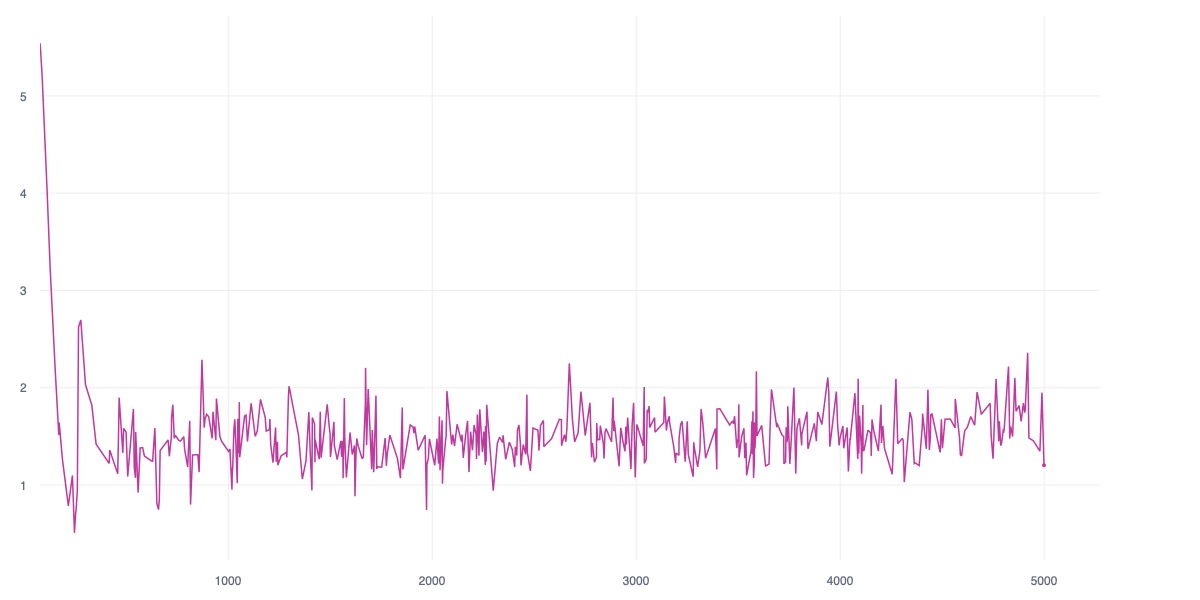}
    \caption{Highway environment: reward prediction loss of the RSSM over training steps.}
    \label{fig:highway_reward_loss}
\end{figure}


\begin{figure}[p]
    \centering
    \includegraphics[width=0.85\linewidth]{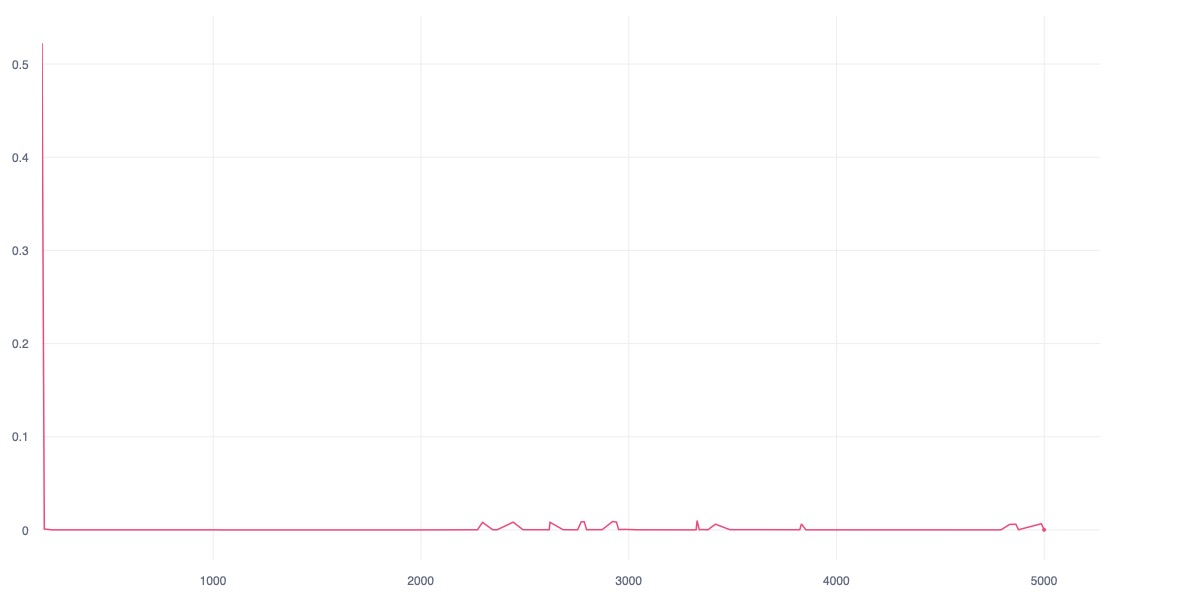}
    \caption{Roundabout environment: continuation loss of the RSSM as a function of training steps.}
    \label{fig:roundabout_cont_loss}
\end{figure}

\begin{figure}[p]
    \centering
    \includegraphics[width=0.85\linewidth]{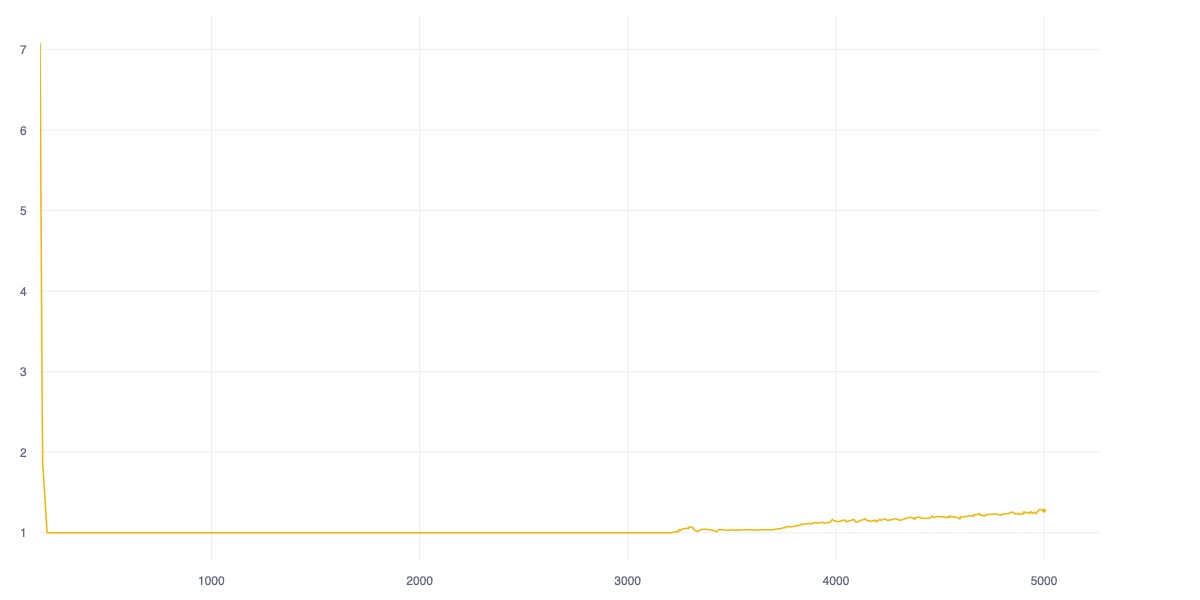}
    \caption{Roundabout environment: dynamics loss of the RSSM as a function of training steps.}
    \label{fig:roundabout_dyn_loss}
\end{figure}

\begin{figure}[p]
    \centering
    \includegraphics[width=0.85\linewidth]{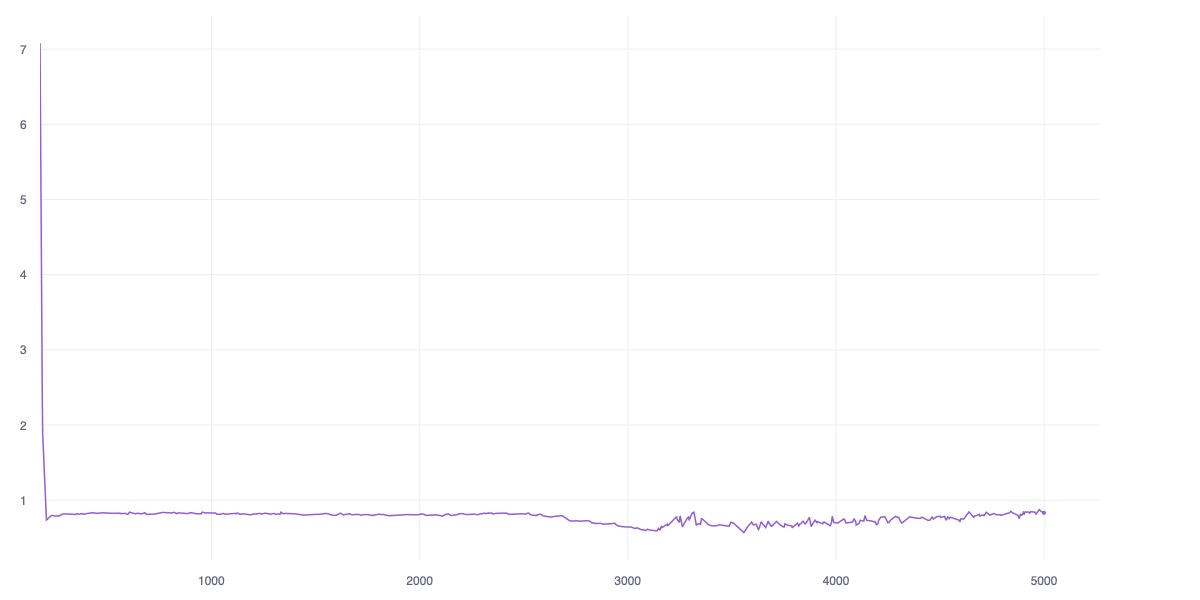}
    \caption{Roundabout environment: KL regularization loss of the RSSM over training steps.}
    \label{fig:roundabout_kl_loss}
\end{figure}

\begin{figure}[p]
    \centering
    \includegraphics[width=0.85\linewidth]{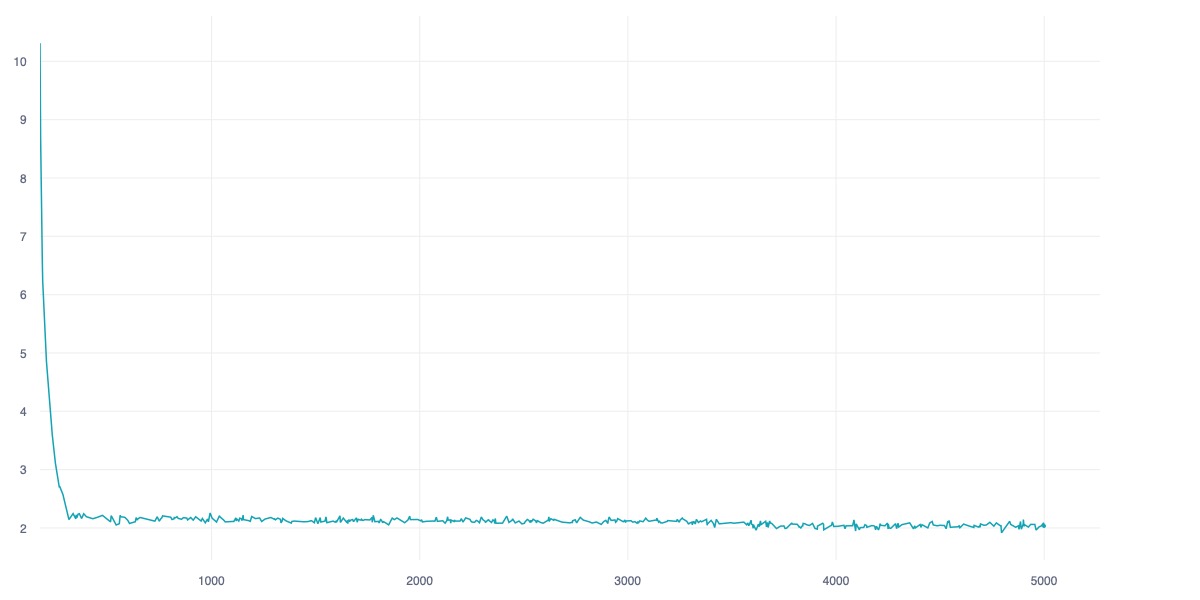}
    \caption{Roundabout environment: overall RSSM model loss over training steps.}
    \label{fig:roundabout_model_loss}
\end{figure}

\begin{figure}[p]
    \centering
    \includegraphics[width=0.85\linewidth]{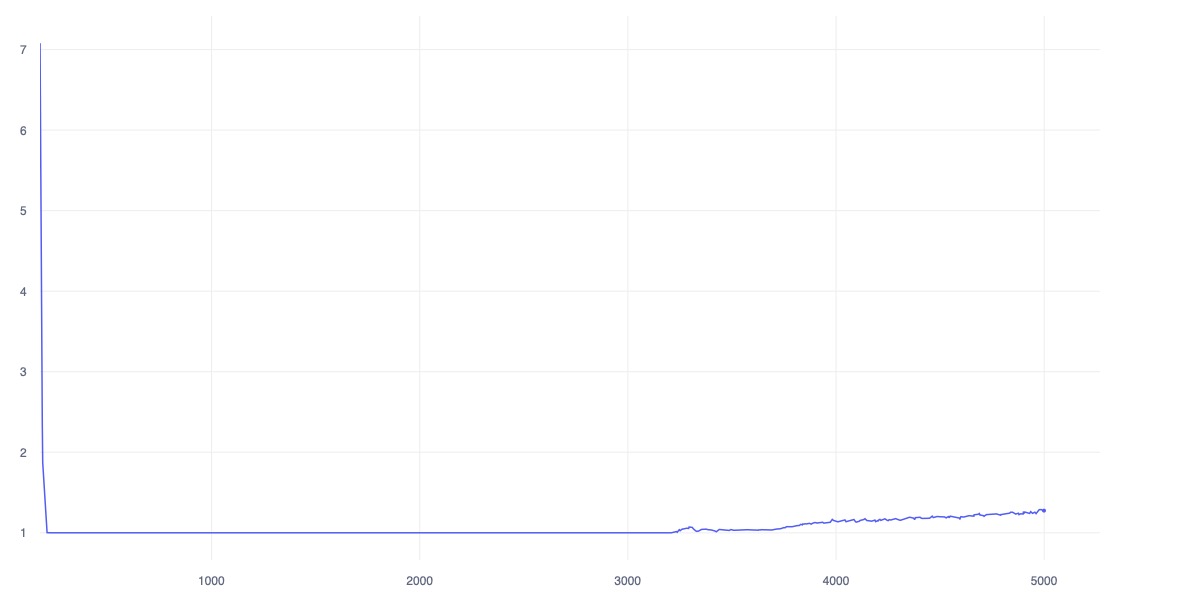}
    \caption{Roundabout environment: representation loss of the RSSM over training steps.}
    \label{fig:roundabout_rep_loss}
\end{figure}

\begin{figure}[p]
    \centering
    \includegraphics[width=0.85\linewidth]{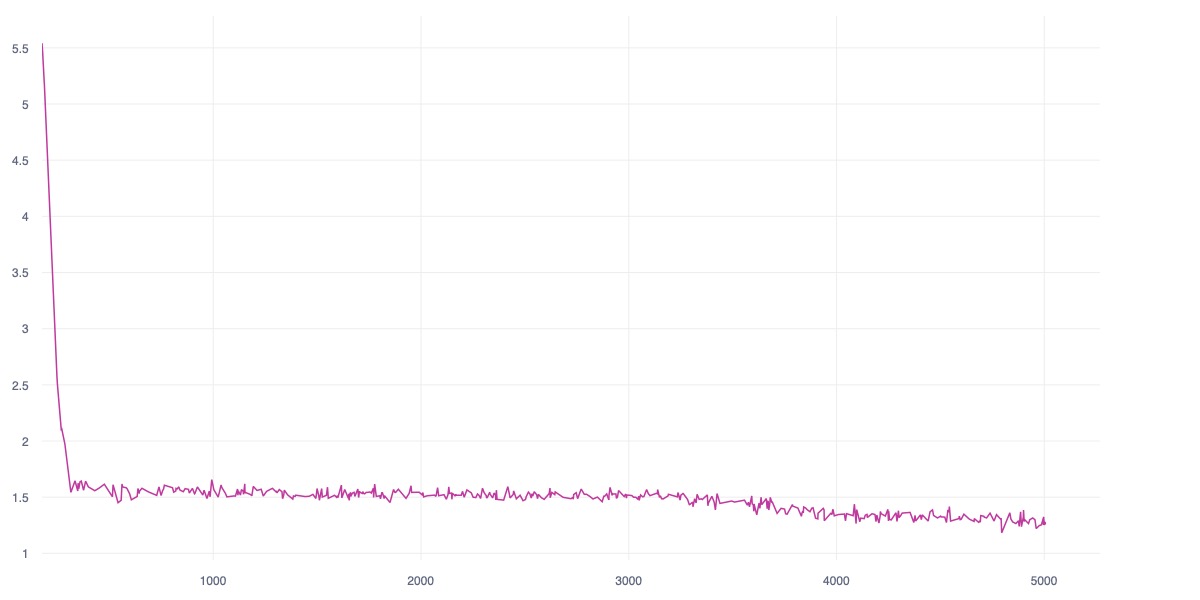}
    \caption{Roundabout environment: reward prediction loss of the RSSM over training steps.}
    \label{fig:roundabout_reward_loss}
\end{figure}


\begin{figure}[p]
    \centering
    \includegraphics[width=0.85\linewidth]{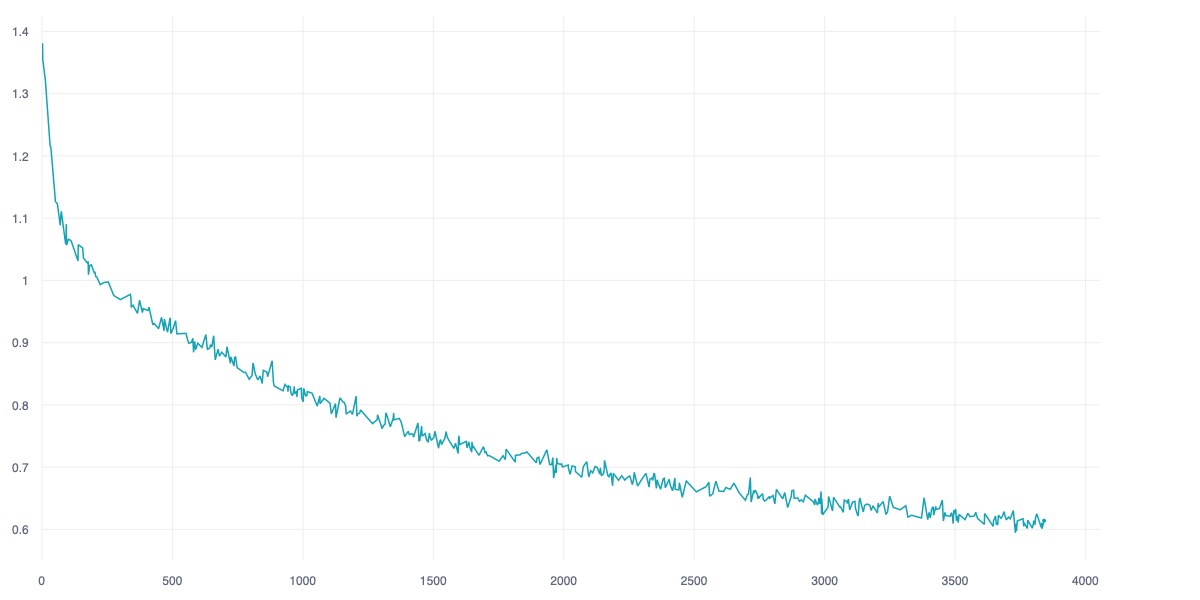}
    \caption{L1 alignment loss between the student spatial predictor and the teacher bottleneck representations in the V-JEPA~2 encoder.}
    \label{fig:encoder_l1_loss}
\end{figure}

\begin{figure}[p]
    \centering
    \includegraphics[width=0.85\linewidth]{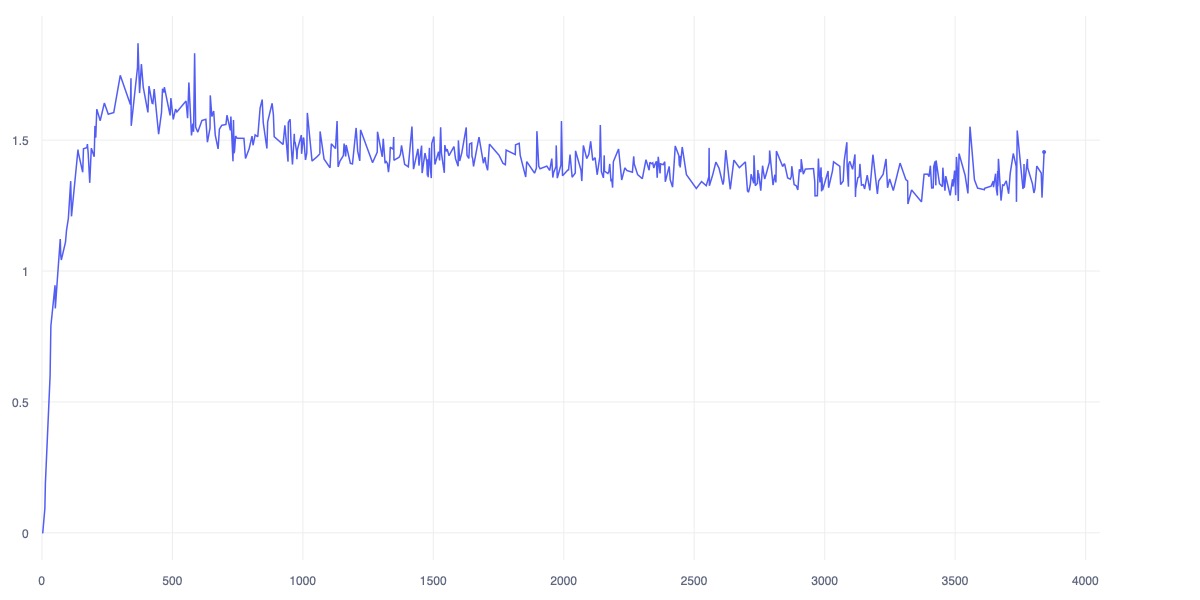}
    \caption{Covariance regularization loss between the student spatial predictor and the teacher bottleneck representations in the V-JEPA~2 encoder.}
    \label{fig:encoder_cov_reg}
\end{figure}

\begin{figure}[p]
    \centering
    \includegraphics[width=0.85\linewidth]{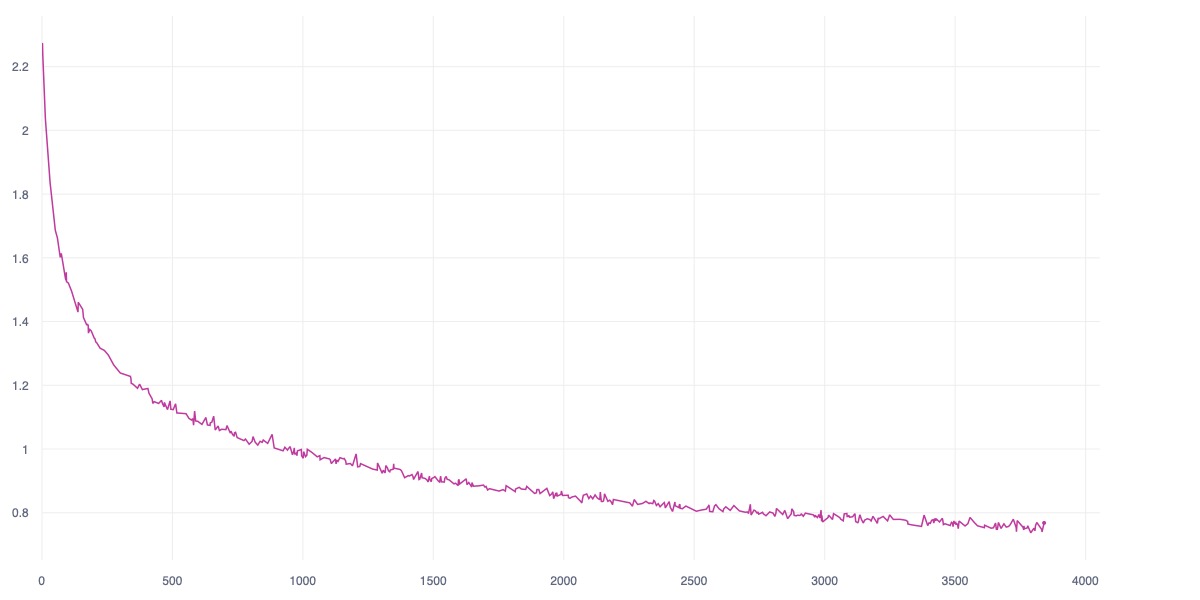}
    \caption{Total encoder training loss for embedding prediction between the student spatial predictor and the teacher bottleneck representations in the V-JEPA~2 encoder.}
    \label{fig:encoder_tot_loss}
\end{figure}

\begin{figure}[p]
    \centering
    \includegraphics[width=0.85\linewidth]{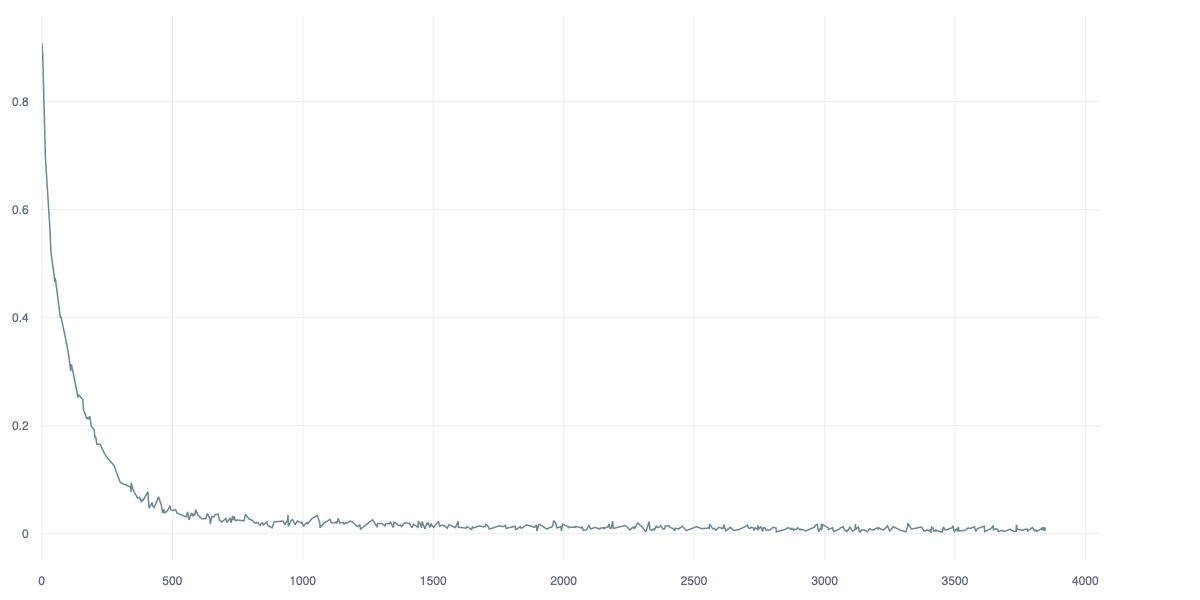}
    \caption{Variance regularization loss applied to the predicted embeddings between the student spatial predictor and the teacher bottleneck representations in the V-JEPA~2 encoder.}
    \label{fig:encoder_var_reg}
\end{figure}

\end{document}